\documentclass[11pt,a4paper]{article}
\usepackage{acl}
\usepackage{times,latexsym}
\usepackage[T1]{fontenc}
\usepackage{microtype}

\usepackage{amsmath}
\usepackage{amsthm}
\usepackage{amsfonts}
\usepackage{amssymb}
\usepackage{mathtools}
\usepackage{comment}

\usepackage{tikz}
\usetikzlibrary{bayesnet}
\usetikzlibrary{arrows}
\usetikzlibrary{decorations}

\usepackage{graphicx}
\usepackage{booktabs}
\usepackage{tabularx}
\usepackage{tabulary}
\usepackage{colortbl}
\usepackage{xspace}
\usepackage{xcolor}
\newcolumntype{Y}{>{\centering\arraybackslash}X}
  \newcolumntype{P}{>{\raggedleft\arraybackslash}X}
\usepackage{multirow}
\usepackage{stfloats}

\usepackage{algorithm}
\usepackage[noend]{algorithmic}

\usepackage[noabbrev,capitalise]{cleveref}

\usepackage{caption}
\usepackage{subcaption}
\usepackage{makecell}
\usepackage{placeins}

\usepackage{amsmath,amsfonts,bm}

\def\eqref#1{equation~\ref{#1}}

\def\1{\bm{1}}

\DeclareMathAlphabet{\mathsfit}{\encodingdefault}{\sfdefault}{m}{sl}
\SetMathAlphabet{\mathsfit}{bold}{\encodingdefault}{\sfdefault}{bx}{n}

\DeclareMathOperator*{\argmax}{arg\,max}

\usepackage{todonotes}
\makeatletter
\newcommand*\iftodonotes{\if@todonotes@disabled\expandafter\@secondoftwo\else\expandafter\@firstoftwo\fi}
\makeatother

\definecolor{edolime}{rgb}{0.9,1,0.3}

\usepackage{xspace}

\title{Probing the Emergence of Cross-lingual Alignment during LLM Training}

\newcommand{\edin}{\epsilon}
\newcommand{\cam}{\kappa}

\author{Hetong Wang$^{\edin}$ \qquad Pasquale Minervini$^{\edin}$ \qquad Edoardo M. Ponti$^{\edin,\cam}$ \\
$^{\edin}$University of Edinburgh \qquad $^{\cam}$University of Cambridge\\
\texttt{H.Wang-197@sms.ed.ac.uk} \\
}

\date{}

\begin{document}
\maketitle

\begin{abstract}
Multilingual Large Language Models (LLMs) achieve remarkable levels of zero-shot cross-lingual transfer performance.
We speculate that this is predicated on their ability to align languages without explicit supervision from parallel sentences.
While representations of translationally equivalent sentences in different languages are known to be similar \emph{after convergence}, however, it remains unclear how such cross-lingual alignment emerges \emph{during pre-training} of LLMs. 
Our study leverages intrinsic probing techniques, which identify which subsets of neurons encode linguistic features, to correlate the degree of cross-lingual neuron overlap with the zero-shot cross-lingual transfer performance for a given model.
In particular, we rely on checkpoints of BLOOM, a multilingual autoregressive LLM, across different training steps and model scales.
We observe a high correlation between neuron overlap and downstream performance, which supports our hypothesis on the conditions leading to effective cross-lingual transfer.
Interestingly, we also detect a degradation of both implicit alignment and multilingual abilities in certain phases of the pre-training process, providing new insights into the multilingual pretraining dynamics.\footnote{Our code is available at: \url{https://github.com/ErikaaWang/probing-multilingual-dynamics}}
\end{abstract}

\section{Introduction}
\label{sec:Intro}

Language Models (LMs) pre-trained on unlabelled multilingual texts show remarkable performance in zero-shot cross-lingual transfer \cite{workshop2023bloom,  xue2021mt5, conneau-etal-2020-unsupervised}.
In fact, fine-tuning a multilingual LM on annotated data for a downstream task in a source language allows it to perform inference in other target languages, too---although often with varying degrees of degradation \cite{pires-etal-2019-multilingual, wu-dredze-2019-beto, libovický2019languageneutral, wu-dredze-2020-languages}.
Surprisingly, this occurs even when the vocabularies of two languages have a null intersection, i.e., no tokens are shared \citep{artetxe-etal-2020-cross}.
Similarly, if the model scale is sufficiently large, LLMs are able to perform cross-lingual transfer through in-context learning with few examples in the source language \cite{lin-etal-2022-shot}.

This implies that LMs can implicitly align lexica and grammar between languages even in the absence of explicit parallel data. To explain this ability, previous work showed that multilingual LMs can encode texts from different languages into language-agnostic representations \cite[\textit{inter alia}]{muller-etal-2021-first} and that grammatical functions are encoded in the same subsets of neurons \citep{stanczak-etal-2022-neurons}. Nevertheless, the existing literature mainly examined the final model upon convergence. Thus, they fail to explain how cross-lingual alignment \textit{emerges} during self-supervised pre-training and how this impacts zero-shot cross-lingual transfer in downstream tasks.

\begin{figure*}
    \centering
    \includegraphics[width=\textwidth]{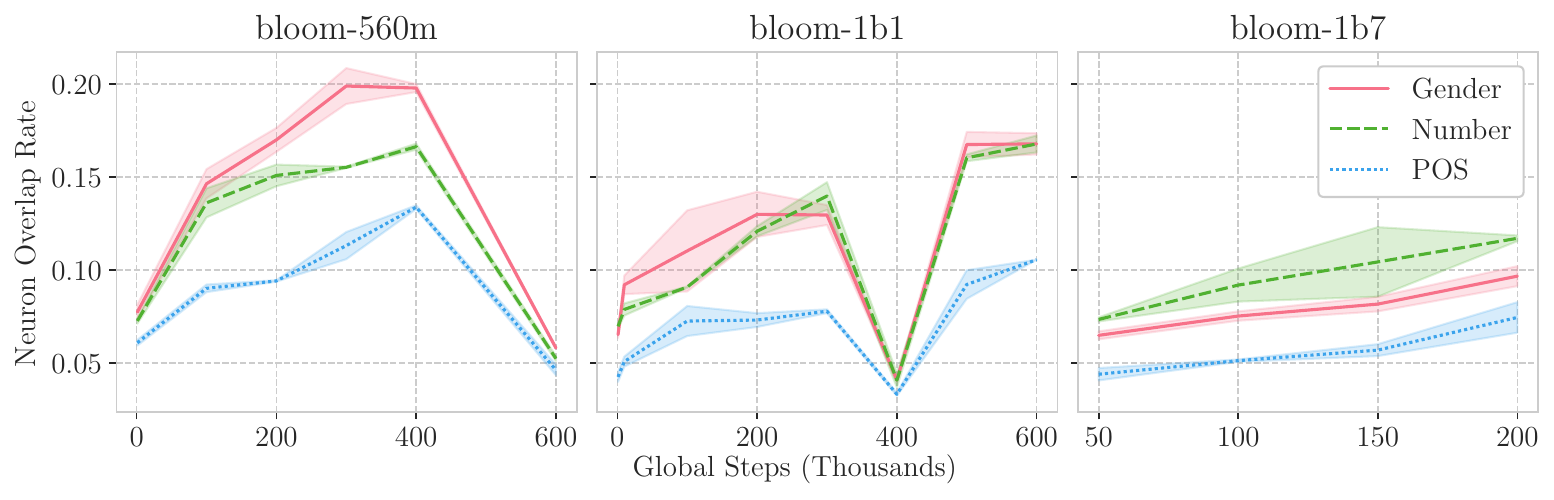}
    \caption{Neuron overlap rates (averaged across pairs of languages) across pre-training steps. Colours and line styles identify selected morphosyntactic categories. The three plots correspond to different model scales.}
    \label{fig:alignment}
\end{figure*}

Hence, our study aims to explore the dynamics of cross-lingual alignment throughout pre-training, discovering trends such as those shown in \cref{fig:alignment}. First, we adopt a reliable intrinsic metric for cross-lingual alignment, namely the extent to which morphosyntactic features (e.g., \textit{Number} for nouns or \textit{Tense} for verbs) tend to activate the same subnetwork within LMs. This implies that the more two languages are aligned, the higher the overlap of the subsets of neurons encoding their information.

We speculate that the degree of alignment tends to increase during pre-training, and that this facilitates the emergence of zero-shot transfer capabilities.
To corroborate this hypothesis, we calculate the correlation between intrinsic metrics of alignment and cross-lingual downstream task performance. To identify neuron overlap at different stages of pre-training, we rely on intrinsic probing \cite{intrinsic_probing}.
Specifically, we probe several checkpoints of BLOOM \cite{workshop2023bloom}, a prominent multilingual LM. This also allows us to compare the emergence of cross-lingual alignment at different model scales, from small (560m) to medium-sized (1.7B) LMs. 

To measure the relation between implicit alignment and downstream performance, we evaluate the zero-shot cross-lingual transfer ability of these checkpoint models on part-of-speech tagging in 11 languages from Universal Dependencies \citep[UD;][]{nivre-etal-2017-universal} and natural language inference in 7 languages \citep[XNLI;][]{conneau-etal-2018-xnli}.
We find a statistically significant, strong correlation between neuron overlap and downstream performance across all model scales. Furthermore, we report a somewhat unexpected finding: both metrics do not grow monotonically during pre-training; rather, they may experience severe drops both in the middle and at the end of pre-training in smaller model scales.

\section{Intrinsic Probing}
\label{sec:method}

We first aim to identify the subnetworks that each language activates within LLMs. To this end,
we employ the latent variable model proposed by \citet{torroba-hennigen-etal-2020-intrinsic} for intrinsic probing, which can identify the subset of specific dimensions within a representation that encodes the information for a particular linguistic feature. 
Formally, given a dataset \(\mathcal{D} = \{(\pi^{(n)}, \bm{h}^{(n)})\}^{N}_{n=1}\), where \(\bm{h}^{(n)} \in \mathbb{R}^d\) are $d$-dimensional embeddings and \(\pi^{(n)} \in \Pi\) are labels that belong to an inventory for a particular linguistic feature (e.g., a part of speech or a morphosyntactic category), our goal is to probe representations \(\bm{h}\), with a total neuron set of \(D = \{1, \dots ,d\}\),
to identify the subset \(C^\star \subseteq D\) that contains the \(k\) most informative neurons with respect to the linguistic feature \(\Pi\). %
For example, the labels \(\Pi = \{ \text{Singular}, \text{Plural} \}\) are associated with the morphosyntactic category of \emph{Number}.
In our setup, we extract hidden representations \(\bm{h}\) from BLOOM\textsubscript{560m}, BLOOM\textsubscript{1b1} and BLOOM\textsubscript{1b7}. Thus, \(d \in \{ 1024, 1536, 2048 \}\), respectively.

Since we are interested in probing the subset of most informative neurons $C$, we introduce a latent variable \(C \subseteq D\) in the probe $p_{\bm{\theta}} \left( \pi \mid \bm{h} \right)$:
\begin{equation}\label{equ:probe}
\begin{split}
    p_{\bm{\theta}} \left( \pi \mid \bm{h} \right) & = \sum_{C\subseteq D} p_{\bm{\theta}}\left(\pi, C \mid \bm{h}\right) \\ & = \sum_{C\subseteq D} p_{\bm{\theta}}\left(\pi \mid \bm{h}, C\right) p(C),
\end{split}
\end{equation}
where \(\bm{\theta}\) are the parameters of the probe.
Following the optimal settings in \citet{stanczak-etal-2022-neurons}, we choose a uniform distribution for the prior \(p(C)\).

To estimate the parameters \(\bm{\theta}\), directly optimising its log-likelihood in \cref{equ:probe} is intractable since it requires marginalising over all possible $k$-sized subsets \(C\) of \(D\), which grow as $\binom{d}{k}$. Thus, we optimise its variational lower bound instead:
\begin{align}
    &\mathcal{L}(\bm{\theta})  = \sum_{n=1}^N \log \sum_{C\subseteq D} p_{\bm{\theta}}\left(\pi^{(n)}, C \mid \bm{h}^{(n)}\right) \label{equ:objective} \\
    &\geq \sum_{n=1}^{N}  \left( \mathop{\mathbb{E}}_{C \sim q_{\phi}} \left[ \log p_{\bm{\theta}}(\pi^{(n)}, C \mid \bm{h}^{(n)}) \right] + H \left(q_{\phi}\right) \right), \notag
\end{align}
\noindent where \(H \left(q_{\phi}\right)\) is the entropy of \(q_{\phi}\), a variational distribution over \(C\) parameterised by \(\phi\).\footnote{The full derivation is available in \citet{intrinsic_probing}. }
\citet{intrinsic_probing} showed that the Poisson sampling is a practically efficient sampling scheme for \(q_{\phi}(C)\), in which each dimension is considered to be independently sampled from a Bernoulli distribution. Therefore, we opt for the Poisson sampling scheme in our setup. 

After having trained the probe model $p_{\bm{\theta}} \left( \pi \mid \bm{h} \right)$ on the morphosyntactic category \(\Pi\), we determine the most informative subset \(C^\star\) by maximising the posterior: 
\begin{equation}\label{equ:select}
    C^\star = \argmax_{C\subseteq D, |C| = k} \sum_{n=1}^{N} \log p\left( \pi^{(n)} \mid \bm{h}^{(n)}_{C} \right)
\end{equation}
where \(\bm{h}_{C}\) is the masked sub-vector of \(\bm{h}\) that contains only dimensions in \(C\).
Since the above combinatorial optimisation problem is intractable in practice, we use a greedy search method for selecting neurons \(1\) to \(k\).

\section{Experimental Setup}

\paragraph{Models.} We conduct the following experiments on BLOOM \cite{workshop2023bloom}, an open-access autoregressive multilingual LM that is jointly trained on data from 46 natural languages and 13 programming languages.
The list of covered languages and their ISO codes is available in Appendix~\ref{apdx:languages}.
In particular, we consider three model sizes: 560m, 1b1, and 1b7, with 6, 8, and 4 valid intermediate model checkpoints, respectively.\footnote{\url{https://huggingface.co/bigscience/bloom-intermediate}. We discovered that the released checkpoints of 1) BLOOM\textsubscript{560m} at steps 10k and 500k, 2) BLOOM\textsubscript{1b7} at steps 1k and 10k, and 3) BLOOM\textsubscript{1b7} at steps 250k and 300k are duplicate model pairs. Thus, we remove these invalid models from the checkpoint collection to ensure reliability.}
Both the checkpoints of BLOOM\textsubscript{560m} and BLOOM\textsubscript{1b1} spread evenly from \textit{1k} to \textit{600k} global training steps, and BLOOM\textsubscript{1b7} ranges from \textit{1k} to \textit{300k}, where the global batch size is increased from 256 to 512. Moreover, BLOOM models with different sizes are trained on an equivalent amount of tokens---which is around 341 billion from the ROOTS corpus \cite{laurençon2023roots_bigscience}---and share the same tokenizer. All these configuration designs allow us to consistently study their training trajectories across scales. 

We studied the cross-lingual ability of BLOOM through two metrics: (i) neuron overlap between languages (\cref{exp:intrinsic}); (ii) zero-shot cross-lingual transfer performance on XNLI and on POS tagging (\cref{exp:ability_measure}), which require multilingual semantic and syntactic knowledge, respectively. In the next section, we will report how these metrics change across pre-training steps and how they correlate with each other. %

\subsection{Intrinsic Probing}\label{exp:intrinsic}

\paragraph{Data.} In order to collect the dataset \(\mathcal{D}\) mentioned in Section \ref{sec:method}, we take advantage of annotated sentences from Universal Dependencies (UD) treebanks v2.1 \cite{nivre-etal-2017-universal} from 13 languages. 
The UD labels are first mapped to the UniMorph Schema \cite{kirov-etal-2018-unimorph} by the converter proposed by \citet{mccarthy-etal-2018-marrying} to ensure a unified label scheme across languages. 
Then, we compute the contextual representations of each word by BLOOM at selected layers. If words are tokenised into subwords, we represent them as the average of their token embeddings, following \citet{vulic-etal-2020-probing}. %
After that, these embedding--label pairs are grouped by linguistic feature (part of speech, number, gender, etc.) and randomly shuffled. Finally, they are split into train, validation, and test sets so that words with the same lemma (e.g. \textit{eat} and \textit{ate}, \textit{employ} and \textit{employer}) appear in the same set. This avoids trivial memorisation of lemma-related information during probe training. 
Additionally, words with lemmas occurring less than 20 times in a split are discarded. This procedure finally results in a batch of datasets \(\mathcal{D}\), each corresponding to a particular morphosyntactic feature, a specific language, and a specific layer depth from which representations are extracted. The available language--feature pairs are listed in Appendix \ref{apdx:language-category-list}. 

\paragraph{Training. } An individual probe is trained for each dataset \(\mathcal{D}\) to identify the neurons that encode most information for the corresponding morphosyntactic feature in a specific language. The probes are trained on the training set with the objective function in \cref{equ:objective}. The probe with parameters \(\bm{\theta}\) is a linear projection followed by a softmax:
\begin{equation}
    p\left(\pi^{(n)} \mid \bm{h}^{(n)}_{C}\right) = \text{softmax}\left(W\bm{h}^{(n)}_{C}\right)
\end{equation}
where $W \in \mathbb{R}^{|\Pi| \times d}$.
After training the probes, neuron sets \(C^\star\) are chosen greedily on validation sets by \cref{equ:select}, where \(\bm{h}^{(n)}_{C}\) is performed by masking all non-chosen dimensions as zero. Based on the results presented in \citet{stanczak-etal-2022-neurons}, we set \(k=50\) as a compromise between performance and computational efficiency. 

\paragraph{Cross-lingual Alignment Metric.} We use the average overlap rate over all possible language pairs for a specific morphosyntactic feature as the metric for cross-lingual alignment. Specifically, we compute the overlap rate of the 50 dimensions probed for each possible language pair where a morphosyntactic feature is expressed, as listed in Appendix \ref{apdx:language-category-list}. Each category will result in an overlap rate matrix. Examples are displayed as heatmaps in Appendix \ref{apdx:heatmap_example}. 

\begin{figure}[t]
    \centering
    \includegraphics[width=0.5\textwidth]{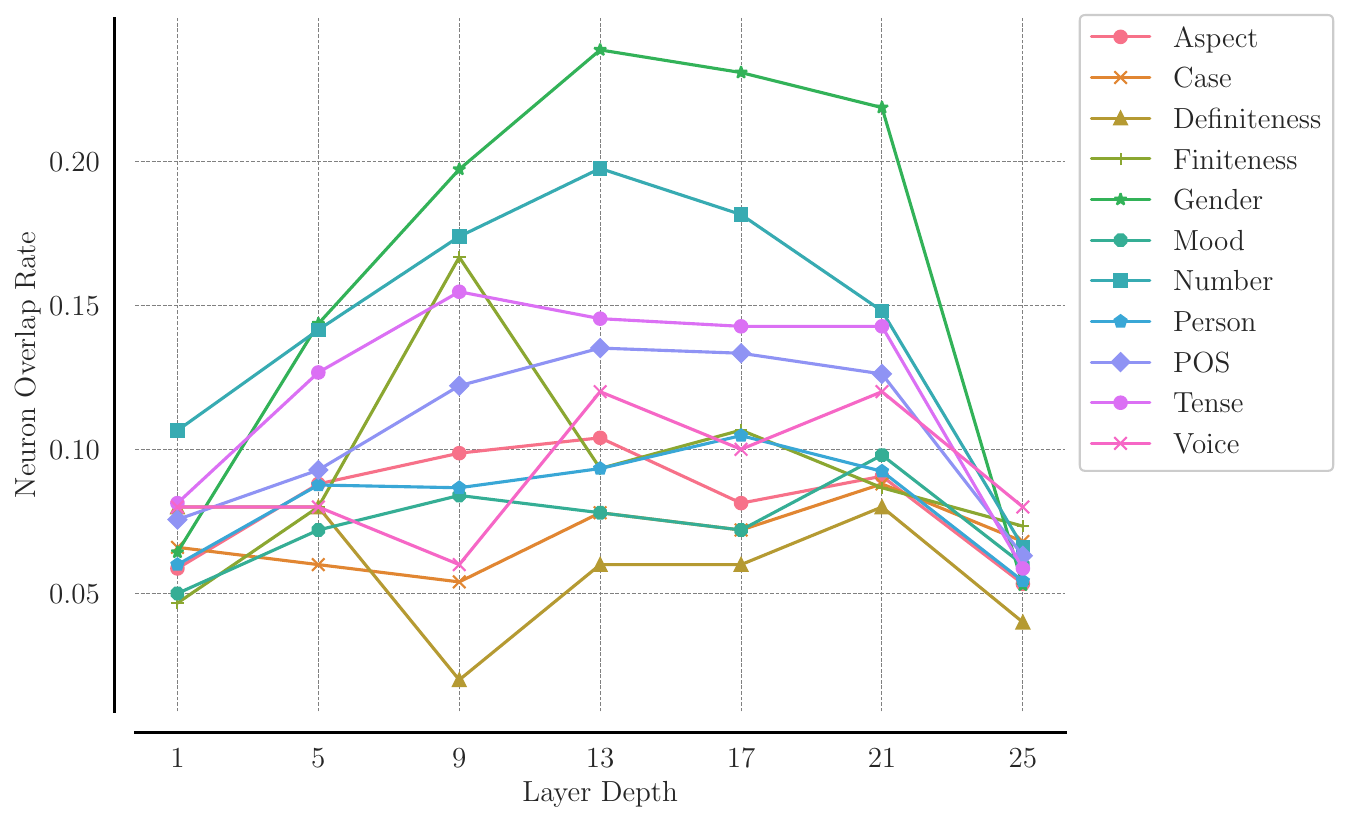}
    \caption{The extent of alignment through layers in the converged BLOOM\textsubscript{560m}. }
    \label{fig:ovlp_layer}
\end{figure} 

\paragraph{Selection of Layers and Linguistic Features.}
To focus on selected layers and linguistic features,
we first exhaustively examine 7 equally distributed layers of the converged BLOOM\textsubscript{560m} on 11 morphosyntactic features by intrinsic probing. \Cref{fig:ovlp_layer} illustrates the extent of cross-lingual alignment throughout different layer depths. 
By comparing the average pairwise overlap rate among linguistic features, the neurons that encode information about \textit{Number} and \textit{Gender} overlap the most, peaking at layers 13 and 17 out of 25. 
Other linguistic features, such as \textit{Case}, \textit{Mood} and \textit{Tense} display a more even trend throughout layers, amounting to 7\%, 8\% and 15\% on average. 
There are also fluctuations around layer 9, where the alignment of \textit{Finiteness} jumps to a peak, while \textit{Voice} and \textit{Definiteness} decrease sharply. 

Incidentally, we observe a drastic decrease in overlap rates at the last hidden layer across all 11 features. 
This phenomenon contrasts with the trend observed in encoder-only models such as m-BERT and XLM-R \citep{intrinsic_probing, stanczak-etal-2022-neurons}, where a significant overlap is observed at the output layer. 
This difference is intuitive, as it aligns with the training objective of different model architectures: encoder-only models are optimised on a Masked Language Modelling objective to replicate the original token, whereas autoregressive models, such as BLOOM, are trained on Next Token Prediction objectives.

Based on the aforementioned results, we select the features \textit{Number} and \textit{Gender}, as well as layers 13 and 17, for our experiments on checkpoints, as they are overall the most informative.
Additionally, we include \textit{POS} tags as a linguistic feature, as it provides the largest language coverage. 
The other two model scales (1b1 and 1b7) also have the same amount of total layers (25), which allows us to adopt an identical policy in layer selection.

\begin{figure*}
    \centering
    \includegraphics[width=\textwidth]{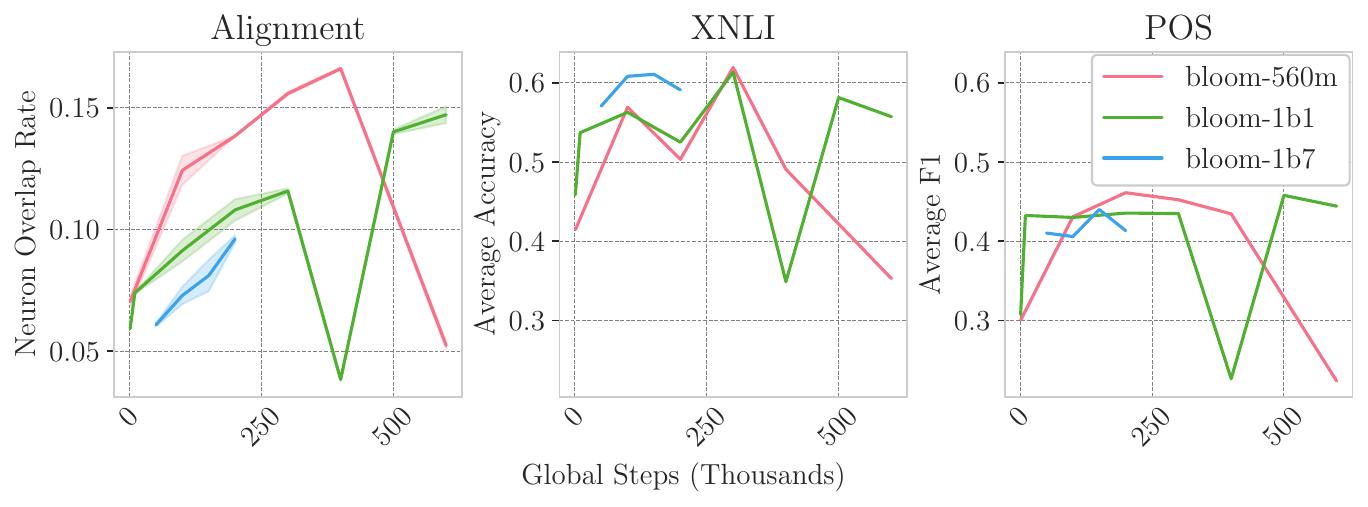}
    \caption{\textbf{Left}: The trend of neuron overlap rates (averaged between layers 13 and 17) throughout training. Line colours indicate different model scales. \textbf{Centre} and \textbf{Right}: Average zero-shot cross-lingual transfer performance across target languages on XNLI and POS tagging.}\label{fig:average_correlation}
\end{figure*}

\subsection{Cross-lingual Transfer Evaluation}\label{exp:ability_measure}
We evaluate the zero-shot cross-lingual transfer ability of the checkpoint collections of BLOOM by two kinds of downstream tasks. Similar to \citet{hu2020xtreme}, we focus on single-source transfer: the annotated training and validation data is provided only in the source language, English. The trained model is directly tested on target languages. We opt for (i) the XNLI dataset as a sentence classification task, and (ii) POS tagging as a structured prediction task. Both are part of widespread multilingual benchmarks such as XTREME \citep{hu2020xtreme}: we follow the same data splits. 

\begin{itemize}
    \item \textbf{XNLI } The Cross-lingual Natural Language Inference dataset, dubbed XNLI \cite{conneau-etal-2018-xnli}, is designed to evaluate the sentence understanding abilities in target languages by determining the relationship between two sentences. The relationships considered are whether the premise \textit{entails}, \textit{contradicts}, or is \textit{neutral} towards the hypothesis. 
    \item \textbf{POS } Part-of-speech tagging data is sourced from UD treebanks \cite{nivre-etal-2017-universal}. These treebanks consist of sentences in a wide range of languages, where each word is annotated with one of the 17 universal POS tags. 
\end{itemize}

\paragraph{Training.} We finetune each checkpoint with the same hyperparameter setting to ensure a fair comparison of their cross-lingual transfer ability.
We use the AdamW~\cite{loshchilov2019adamw} optimiser with a learning rate %
of $2 \times 10^{-5}$.
The models are trained for 5 epochs on XNLI with a training set of 392k samples and 10 epochs on POS tagging with a training set of 21k samples.
We perform model selection based on development set performance, evaluating models every 100 (POS) or 500 (XNLI) steps.
Finally, we evaluate the best finetuned models on the test set of each target language, using accuracy as a metric for XNLI and F1 score for POS tagging. 

Limited by computational resources, we conduct this fine-tuning with qLoRA \cite{dettmers2023qlora}. This first quantises the (frozen) pre-trained LLM to 4-bit and then applies a trainable Low-Rank Adapter \citep[LoRA;][]{hu2021lora}. The adapter only requires around 10\% of the original model parameters, and each model could be fine-tuned on a single 80GB NVIDIA A100 GPU.\footnote{To make the results comparable across scales, we apply the same training setting (including QLoRA) to the three sizes of BLOOM.}

\section{Results}
\label{sec:results}
\subsection{The Dynamics of Alignment}
\label{sec:analyse_dynamics}

The level of cross-lingual alignment during pre-training of BLOOM models is shown in \cref{fig:alignment}. 
The plots exhibit similar trends across the linguistic properties we probed, within the same model scale; however, they differ significantly across scales. This stands in contrast with our initial assumption of a gradual emergence of language-agnostic representations, which would imply a monotonic increase of neuron overlap.
First, we find that the smallest model (BLOOM\textsubscript{560m}) shows the highest overlap during most of the pre-training steps. 
Moreover, we notice a dramatic drop of overlap rates in two model scales, which occurs at around 600k global steps for BLOOM\textsubscript{560m} and a bit earlier at 400k steps for BLOOM\textsubscript{1b1}. 
This drop happens at the end of the pre-training of scale 560m, while BLOOM\textsubscript{1b1} recovers a high rate of neuron overlap in the latter stage of pre-training. 

\begin{table*}[ht]
\centering
\begin{tabular}{cl|ll|ll}
    \toprule
    &  & \multicolumn{2}{c}{\textbf{XNLI}} & \multicolumn{2}{c}{\textbf{POS}} \\
    \cmidrule(lr){3-4}\cmidrule(lr){5-6}
    & & Average & Pairwise & Average & Pairwise \\
    \midrule
    \multirow{3}{*}{\textbf{Pearson ($r$)}} & BLOOM\textsubscript{560m} &  0.808\textsubscript{\textcolor{gray}{0.052}} & \textbf{0.568}\textsubscript{\textcolor{purple}{8.774e-05}} & \textbf{0.940}\textsubscript{\textcolor{orange}{0.005}}& \textbf{0.612}\textsubscript{\textcolor{purple}{3.277e-09}} \\
    & BLOOM\textsubscript{1b1} &  \textbf{0.804}\textsubscript{\textcolor{orange}{0.016}} &\textbf{0.723}\textsubscript{\textcolor{purple}{3.081e-10}} & \textbf{0.831}\textsubscript{\textcolor{orange}{0.011}}&  \textbf{0.638}\textsubscript{\textcolor{purple}{1.204e-12}} \\
    & BLOOM\textsubscript{1b7} &  0.395\textsubscript{\textcolor{gray}{0.605}} &\textbf{0.572}\textsubscript{\textcolor{orange}{0.001}}& 0.258\textsubscript{\textcolor{gray}{0.742}}&  \textbf{0.534}\textsubscript{\textcolor{purple}{2.691e-05}} \\
    \bottomrule
\end{tabular}
\caption{Correlation analysis on average (shown by Fig. \ref{fig:average_correlation}) and pairwise (shown by Fig. \ref{fig:pairwise_correlation_XNLI} and \ref{fig:pairwise_correlation_POS}) overlap rate and zero-shot transfer performance by Pearson coefficient, where the p-values are displayed as subscripts. Colours of p-values indicate \textcolor{orange}{statistical significance ($p<0.05$)}, \textcolor{purple}{high statistical significance ($p<0.001$)} and \textcolor{gray}{no statistical significance ($p \geq 0.05$)}. Coefficients larger than 0.5 with significance under the null hypothesis are bold. }\label{correlation_table}
\end{table*}

While this phenomenon may be an artefact due to the variance of overlap rates or an error in checkpointing, we remark that similar drops were also observed in encoder-only multilingual LMs.
In fact, \citet{blevins-etal-2022-analyzing} also detects a performance degradation point among a series of XLM-R checkpoints when evaluated on dependency relation prediction. This affects both in-language performance and cross-lingual transfer. 
On the other hand, no similar phenomenon occurs when probing monolingual models: \citet{liu-etal-2021-probing-across} report that these LMs usually display a steady acquisition of linguistic properties along the pre-training trajectories, retaining high performance maintains after a steep increase at the beginning. 
In contrast, we find that linguistic features are obtained gradually but inconsistently in multilingual models throughout the pre-training process, as shown in \cref{fig:alignment}.

As BLOOM\textsubscript{560m}, BLOOM\textsubscript{1b1}, and BLOOM\textsubscript{1b7} share the training corpus, hyperparameter setting and architecture, they only differ in model scales; however, only the largest scale, BLOOM\textsubscript{1b7}, shows a monotonic growth in neuron overlap.
Thus, the emergence of cross-lingual alignment might follow a scaling law \cite{kaplan2020scaling} only after a certain threshold in model size. This hypothesis is supported by further results discussed in \cref{sec:discussion} and could be verified on larger scales, such as BLOOM\textsubscript{3b} and BLOOM\textsubscript{7b1}.

\subsection{Cross-lingual Transfer Correlation}

\begin{figure*}[t]
    
    \begin{subfigure}[b]{\textwidth}
    \centering
    \includegraphics[width=\textwidth]{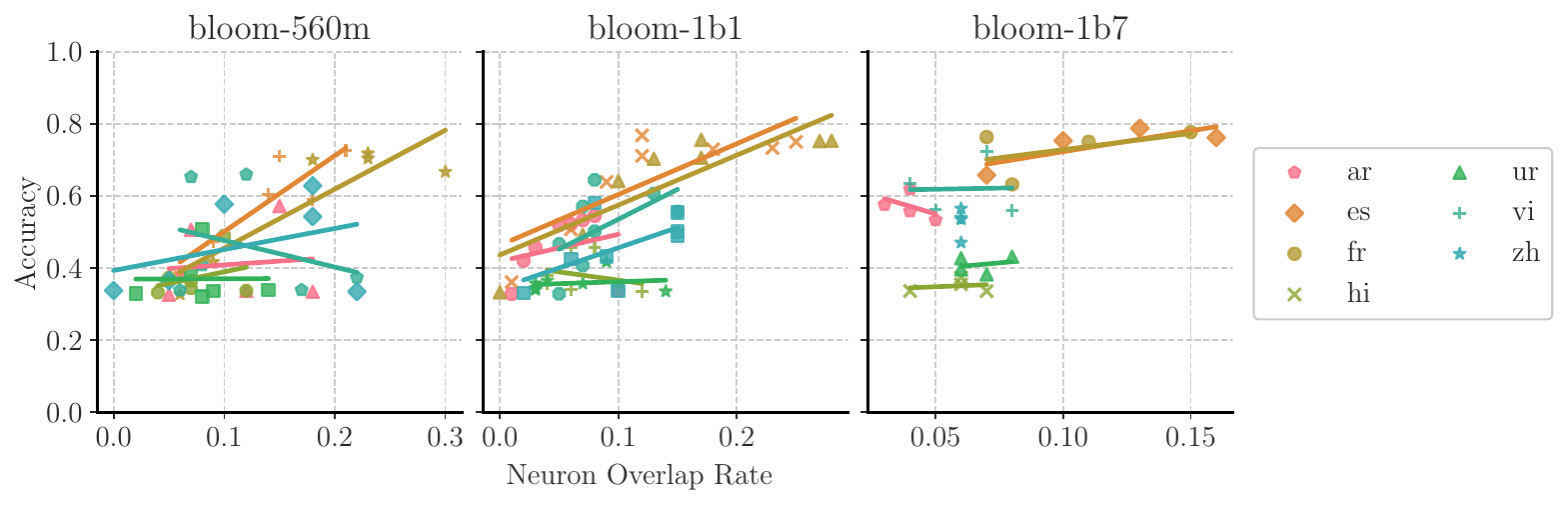}
    \caption{Pairwise cross-lingual correlation on XNLI. }\label{fig:pairwise_correlation_XNLI}
    \end{subfigure}
    
    \begin{subfigure}[b]{\textwidth}
    \centering
    \includegraphics[width=\textwidth]{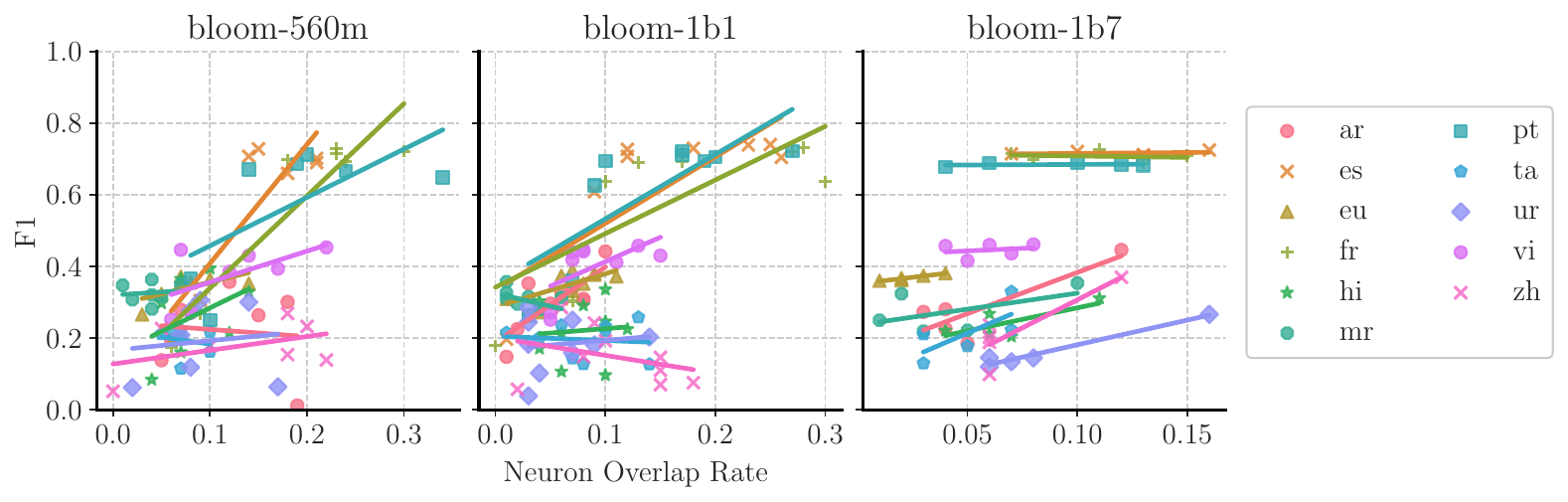}
    \caption{Pairwise cross-lingual correlation on POS. }\label{fig:pairwise_correlation_POS}
    \end{subfigure}
    
    \caption{Neuron overlap rate, which measures the extent of cross-lingual alignment, plotted against the zero-shot cross-lingual transfer performance on (a) XNLI and (b) POS tagging for all checkpoints. The colours indicate the target languages paired with English.
    The Pearson's correlation coefficients are shown in Table \ref{correlation_table}.}
    \label{fig:overall_correlation}
    
\end{figure*}

We also conduct a correlation analysis between the cross-lingual alignment and the zero-shot transfer performance, illustrated in \cref{fig:average_correlation}. 
Overall, the zero-shot cross-lingual transfer ability of BLOOM shows a strong correspondence with the neuron overlap throughout pre-training, within each model size. This observation holds true also when considering each target language individually, rather than the cross-lingual average, as shown in \cref{fig:pairwise_correlation_XNLI} and \cref{fig:pairwise_correlation_POS}. 

We measure the strength of the correlation in terms of Pearson's coefficients $r$ and its statistical significance against the null hypothesis as p-values. As shown in Table \ref{correlation_table}, we compute the correlations on the average (\cref{fig:average_correlation}) and pairwise (\cref{fig:pairwise_correlation_XNLI} and \cref{fig:pairwise_correlation_POS}) neuron overlap for each model size against both XNLI and POS tagging performance. The correlations are noticeably higher overall for pairwise measurements (as opposed to average metrics) and for the two smaller models (560m and 1b1). Nonetheless, the fact that pairwise correlations for all model scales are both strong and significant lends credibility to our claim that neuron overlap is tightly connected with zero-shot cross-lingual transfer abilities. Thus, we verify that multilingual LMs transfer between languages more easily if more shared neurons are aligned while pretraining. %

\subsection{Why does the drop point occur?}\label{sec:discussion}

\begin{figure*}[t]
    \centering
         \begin{subfigure}[b]{0.45\textwidth}
         \centering
         \includegraphics[width=\textwidth]{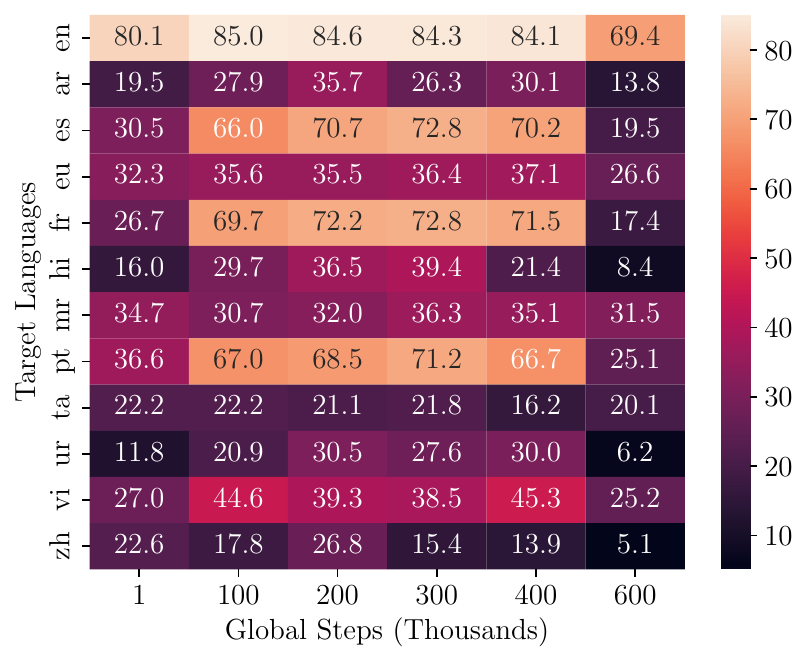}
         \caption{BLOOM\textsubscript{560m} on POS tagging.}
     \end{subfigure}   
     \begin{subfigure}[b]{0.45\textwidth}
         \centering
         \includegraphics[width=\textwidth]{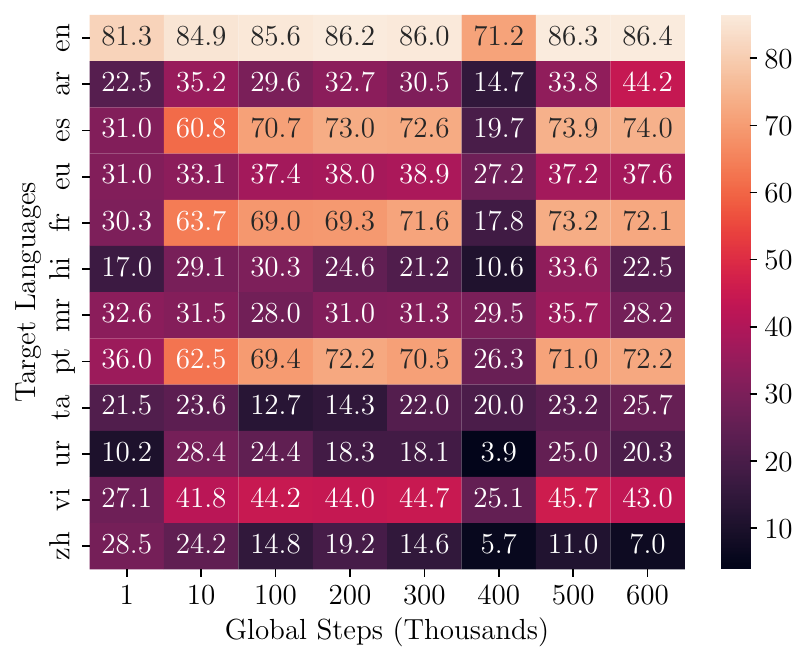}
         \caption{BLOOM\textsubscript{1b1} on POS tagging.}
     \end{subfigure}  

         \begin{subfigure}[b]{0.45\textwidth}
         \centering
         \includegraphics[width=\textwidth]{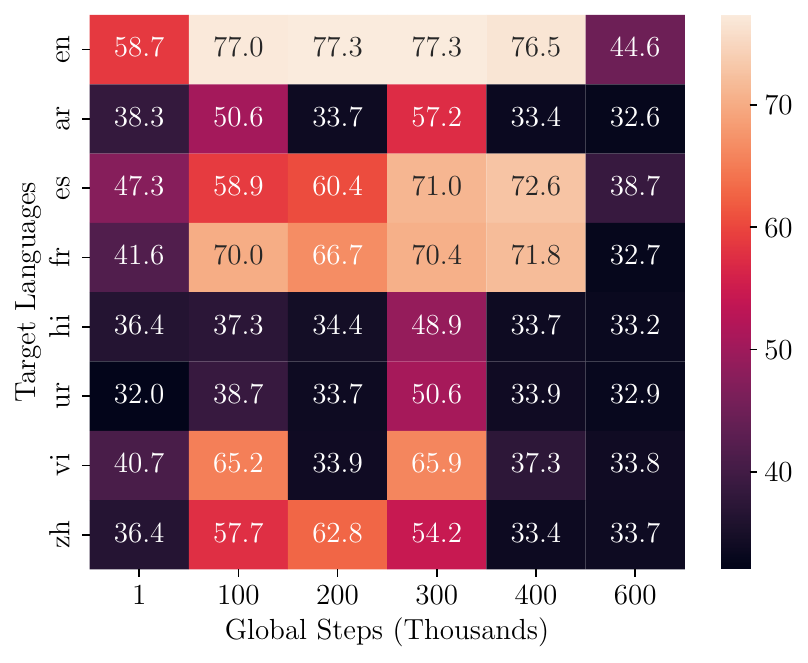}
         \caption{BLOOM\textsubscript{560m} on XNLI.}
     \end{subfigure}   
     \begin{subfigure}[b]{0.45\textwidth}
         \centering
         \includegraphics[width=\textwidth]{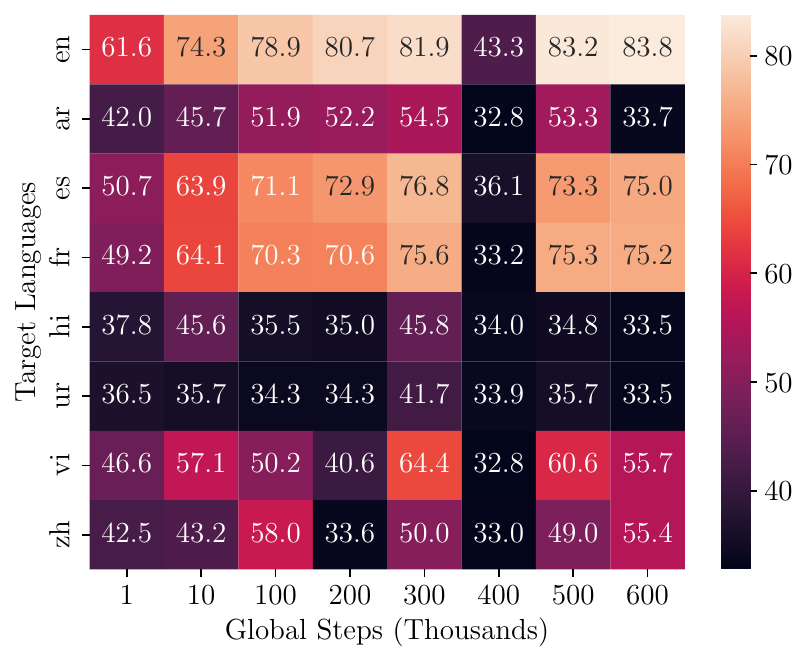}
         \caption{BLOOM\textsubscript{1b1} on XNLI.}
     \end{subfigure}   
    \caption{The zero-shot cross-lingual performance of BLOOM checkpoints of size 560m (left) and 1b1 (right) on POS tagging (top) and XNLI (bottom). }
    \label{fig:body_heatmap_f1}
\end{figure*}

As already mentioned in §\ref{sec:analyse_dynamics}, we observe that an unexpected drop of cross-lingual alignment may appear during the training process, suggesting that multilingual LMs can pass through highly sub-optimal regions in the loss landscape. As depicted by the drop points in \cref{fig:alignment} and \cref{fig:average_correlation}, intermediate models saved on global step \textit{600k} of BLOOM\textsubscript{560m} and \textit{400k} of BLOOM\textsubscript{1b1} have common traits: \textbf{1)} there is nearly no neuron overlap detected in these models; \textbf{2)} These models display weak zero-shot transfer abilities since their performance on target languages are mostly random, as shown in \cref{fig:body_heatmap_f1}; \textbf{3)} These models' performance becomes worse also in-language, showing a performance degradation in the source language English.%

Several works studying linguistic acquisition in LMs across time find that the risk of models falling into bad minima depends on their scale. %
\citet{xia2023training} conduct a study on the training dynamics across scales in monolingual LMs. They find that while all models \textit{decrease} their perplexity for hallucinated texts at the start of training, only large-scale models eventually escape this sub-optimal distribution. \citet{conneau-etal-2020-unsupervised} introduce the \textit{curse of multilinguality}, which refers to the phenomenon that given a fixed number of parameters, continued increase in the number of languages leads to a performance degradation in terms of both monolingual and cross-lingual skills. Consequently, they argue that this bottleneck could be solved by increasing model scales. 

Our experimental results on the dynamics of cross-lingual alignment are in agreement with the aforementioned works. We suggest that cross-lingual alignment follows the same trajectory of learning across scales, similar to what is observed in \citet{Chen2023SuddenDI}. However, smaller scales appear more likely to pass through or fall into suboptimal parameter configurations, which lead to a simultaneous degradation in both in-language and cross-lingual abilities. Our work thus offers a more nuanced perspective on the multilingual abilities of LMs based on learning dynamics, which enriches the received wisdom based on converged models.

\section{Related work}
\label{related_work}

\paragraph{Probing Linguistic Features in Multilingual LMs.} Probing is a prevalent approach for model interpretability, which is used to examine the information encapsulated in the hidden representation of LMs \cite{taktasheva-etal-2021-shaking, papadimitriou-etal-2021-deep}, including multilingual LMs such as m-BERT and XLM \cite{lample2019crosslingual}. Previous work demonstrated that embedding spaces in different languages tend to be isomorphic, and can be better aligned \textit{post-hoc} with the aid of parallel examples or anchor points, which improves zero-shot cross-lingual performance \cite{cao2020multilingual, schuster-etal-2019-cross,  conneau-etal-2020-emerging}. 

\paragraph{Structural Overlap and Generalisations.}
Overlap in neurons (dimensions of hidden representations) or subnetworks of parameters are considered to support generalisation abilities. A direction of research attempts to identify language-specific neural subnetworks, finding that they are topologically similar \cite{foroutan2022discovering} and that their overlap might depend on typological distance \citep{ansell-etal-2022-composable, ansell-etal-2023-distilling}. In addition, \citet{muller-etal-2021-first} detected a high correlation between the similarity of representations and the zero-shot cross-lingual transfer performance in the converged mBERT. All of these works imply a strong correlation between cross-lingual alignment and zero-shot transfer ability, which is further confirmed by our study on the training trajectory of BLOOM across scales.

Recently, \citet{bhaskar2024heuristic} finds that all competing subnetworks within LLMs, which have similar in-domain performance but different out-of-domain generalisation, share a so-called `heuristc core', while \citet{templeton2024scaling} demonstrated that sparse auto-encoders can identify various features---from a specific landmark to code errors---in a production-grade LLM. 
Our work, in conjunction with theirs, provides a reliable framework for explaining model generalisation through the lens of shared neurons.

\paragraph{Knowledge Acquisition during Pre-training. } 
Concurrently, there is a rising interest in understanding the training dynamics of LLMs. 
Works that mainly examine monolingual English models report a steady trend in the acquisition of linguistic knowledge. 
Both \citet{xia2023training} and \citet{choshen-etal-2022-grammar} argue that language acquisition undergoes the same order of phase transitions consistently across model scales, training objectives and random seeds. 
\citet{Chen2023SuddenDI} find that the emergence of syntactic structure in the attention scores of Transformer-based LMs is essential for grammar acquisition in LMs, but does not account for semantic knowledge acquisition.
For multilingual training, \citet{choenni2023languages} examine how data size and language variance affect the performance during fine-tuning. The experiments presented by \citet{blevins-etal-2022-analyzing} are the most reminiscent of our work. They focus on the inconsistency between the emergence of in-language and cross-language abilities for encoder LMs, whereas we study the dynamics of neuron overlaps and the corresponding impact on downstream performance in autoregressive LMs.

\section{Conclusions}
In this paper, we probe a collection of checkpoint models of BLOOM to study the dynamics of multilingual pretraining. By experimenting with three model sizes, we observe that the subset of neurons encoding linguistic features tends to increase their overlap across languages throughout pretraining. Nevertheless, we also detect severe drops that occur at different points in the training process, especially at smaller model scales, instead of a steady increase in the extent of alignment. 

Moreover, we corroborate the hypothesis that the shared neurons are tightly connected with the zero-shot cross-lingual transfer ability of multilingual LLMs: the same sub-networks are activated at inference time and updated during fine-tuning, which contributes to the cross-lingual generalisation ability of LMs. This assumption is further confirmed across model scales by observing a high correlation between neuron overlap and downstream task performance in syntactic and semantic tasks. Hence, our work contributes to understanding how multilingual LMs implicitly align information across languages even in the absence of parallel data.

\section*{Limitations}
Our work focuses on the checkpoints of BLOOM with large intervals in global steps. Thus, our findings on the trend of alignment might be not applicable if zooming in on a particular window of training with finer-grained checkpoint models. Moreover, we consider only autoregressive models with the same objective and training dataset: varying these properties may result in different patterns.

Although many of our findings on the dynamics of cross-lingual alignment align with previous research on encoder Transformers, some aspects of the experimental design  (e.g., selected layers and morphosyntactic categories) are not directly transferable to other architectures or training corpora. %
Moreover, we focus on the alignment of languages seen during pretraining, whereas the generalisation to unseen languages is left for future research.

\section*{Acknowledgements}

This work used resources provided by the Edinburgh Compute and Data Facility (ECDF).\footnote{http://www.ecdf.ed.ac.uk/}

\bibliography{anthology,custom}

\onecolumn
\appendix

\section{Languages List}

\subsection{Target Languages List}\label{apdx:languages}
Based on our hypothesis introduced in \cref{sec:Intro}, we select the set of target languages from the intersection of treebanks in UD v2.1 and the BigScience ROOTS Corpus \citep{laurençon2023roots_bigscience} used for pre-training BLOOM, so that we can perform implicit alignment detection following the procedure described in §\ref{exp:intrinsic}. In addition to this criterion, we further select target languages based on the data availability for experiments on downstream tasks in \cref{exp:ability_measure}. A full list is given below. 
\FloatBarrier
\begin{table*}[hbt!]
\centering
\begin{tabular}{lccccccccccccc}\toprule 
    UD v2.1  & ar & eu & ca & zh & en & fr & hi & mr & pt & es & ta & ur & vi \\\midrule
    §\ref{exp:ability_measure} XNLI & \checkmark & &  &\checkmark &\checkmark & \checkmark &\checkmark &  & &\checkmark &  &\checkmark & \checkmark \\
    §\ref{exp:ability_measure} POS & \checkmark &\checkmark & &\checkmark &\checkmark & \checkmark &\checkmark & \checkmark &\checkmark &\checkmark & \checkmark &\checkmark & \checkmark \\
    \bottomrule
\end{tabular}
\end{table*}
\FloatBarrier

\subsection{Language and Morphosyntactic categories}\label{apdx:language-category-list}
The following table lists the corresponding morphosyntactic categories for the languages we probed. 
\FloatBarrier
\begin{table*}[hbt!]
\resizebox{\textwidth}{!}{
\begin{tabular}{lccccccccccccc} 
\toprule 
Language & \makecell{ISO \\ 639-1 \\ code} & \makecell{ISO \\639-3 \\code} & Aspect & Case & Definiteness & Finiteness & Gender & Mood & Number & Person & POS & Tense & Voice\\ 
\midrule 
Arabic & ar & ara &  \checkmark & \checkmark & \checkmark & & \checkmark &\checkmark &\checkmark & & \checkmark & & \checkmark \\
\cmidrule{4-14}
Basque & eu & eus & \checkmark &\checkmark &\checkmark & &&& \checkmark & & \checkmark & &  \\
\cmidrule{4-14}
Catalan & ca & cat & &&&&\checkmark &\checkmark &\checkmark &\checkmark &\checkmark &\checkmark &\\
\cmidrule{4-14}
Chinese & zh & zho & &&&&&&&&\checkmark &\\
\cmidrule{4-14}
English & en & eng & &&&&&&\checkmark &\checkmark &\checkmark &\checkmark &\\
\cmidrule{4-14}
French & fr & fra & \checkmark & &&&\checkmark && \checkmark &\checkmark &\checkmark &\checkmark &\\
\cmidrule{4-14}
Hindi & hi & hin & \checkmark &\checkmark && \checkmark &\checkmark &\checkmark &\checkmark &\checkmark &\checkmark && \checkmark\\
\cmidrule{4-14}
Marathi & mr & mar & &&&&\checkmark & & \checkmark &\checkmark &\checkmark &\\
\cmidrule{4-14}
Portuguese & pt & por & \checkmark & &&& \checkmark &\checkmark &\checkmark & \checkmark &\checkmark &\checkmark &\\
\cmidrule{4-14}
Spanish & es & spa & \checkmark & &&& \checkmark &\checkmark &\checkmark & \checkmark &\checkmark &\checkmark &\\
\cmidrule{4-14}
Tamil & ta & tam & & \checkmark & & \checkmark &\checkmark & & \checkmark & & \checkmark &\checkmark &\\
\cmidrule{4-14}
Urdu & ur & urd & & \checkmark & & \checkmark & && \checkmark & & \checkmark &\\
\cmidrule{4-14}
Vietnamese & vi & vie & &&&&&&&&\checkmark &\\
\bottomrule
\end{tabular}
}
\end{table*}
\FloatBarrier

\clearpage
\section{Results}
\subsection{Pairwise Overlap Comparison}\label{apdx:heatmap_example}
In this section, we exhibit an exhaustive collection of heatmaps of layer 17 in the converged BLOOM\textsubscript{560m} for all the possible morphosyntactic categories listed in Appendix \ref{apdx:language-category-list}. The orange dot indicates an overlap that is statistically significant under the null hypothesis. 
\begin{figure}[H]
     \centering
     \begin{subfigure}[b]{0.23\textwidth}
         \centering
         \includegraphics[width=\textwidth]{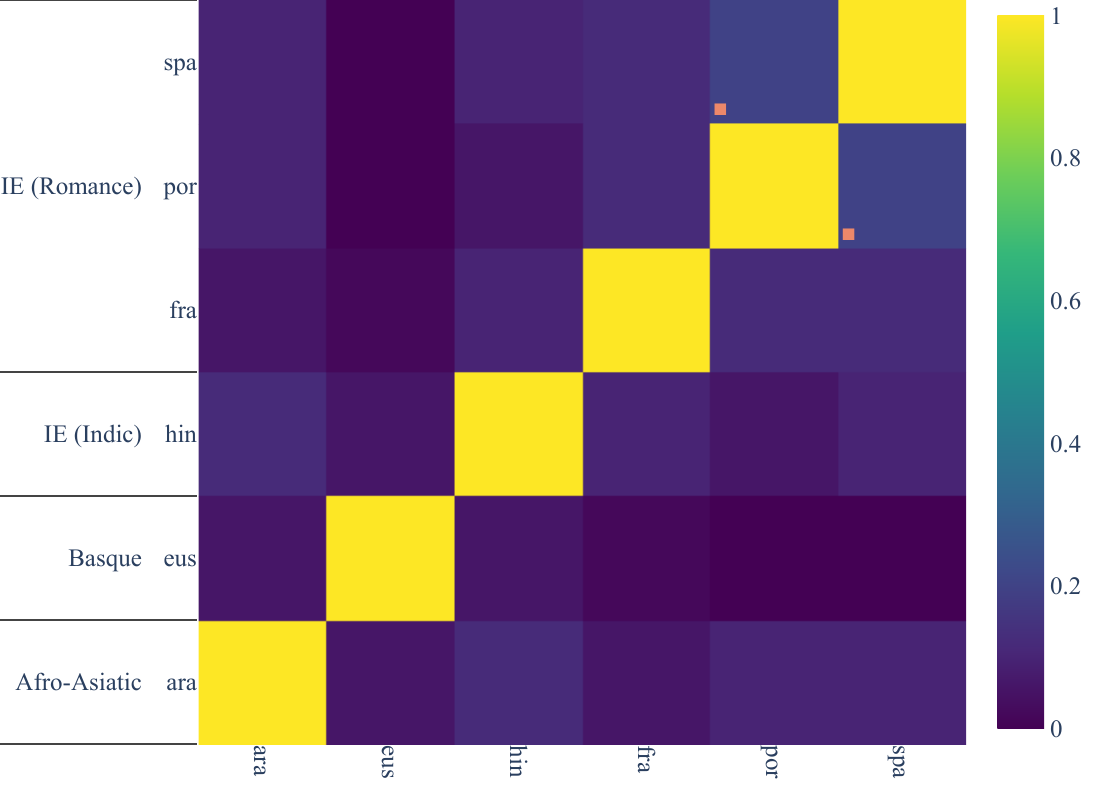}
         \caption{Aspect}
     \end{subfigure}
     \hfill
     \begin{subfigure}[b]{0.23\textwidth}
         \centering
         \includegraphics[width=\textwidth]{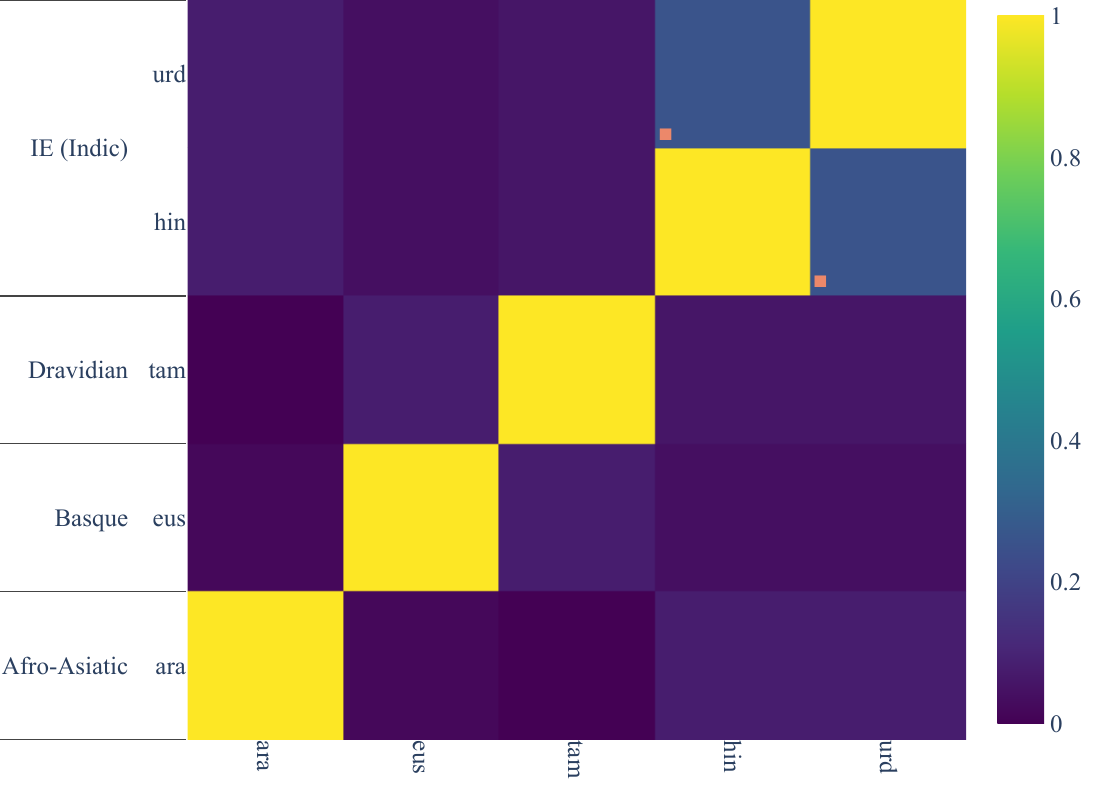}
         \caption{Case}
     \end{subfigure}    
     \begin{subfigure}[b]{0.23\textwidth}
         \centering
         \includegraphics[width=\textwidth]{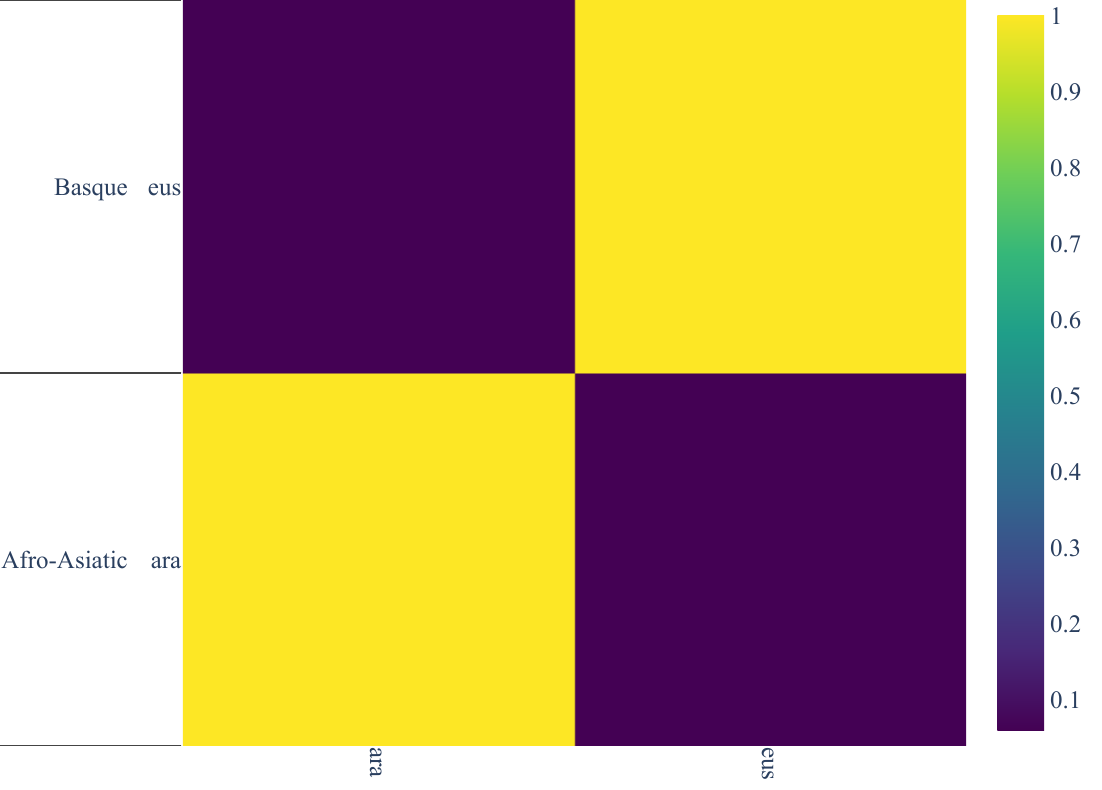}
         \caption{Definiteness}
     \end{subfigure}
     \hfill
     \begin{subfigure}[b]{0.23\textwidth}
         \centering
         \includegraphics[width=\textwidth]{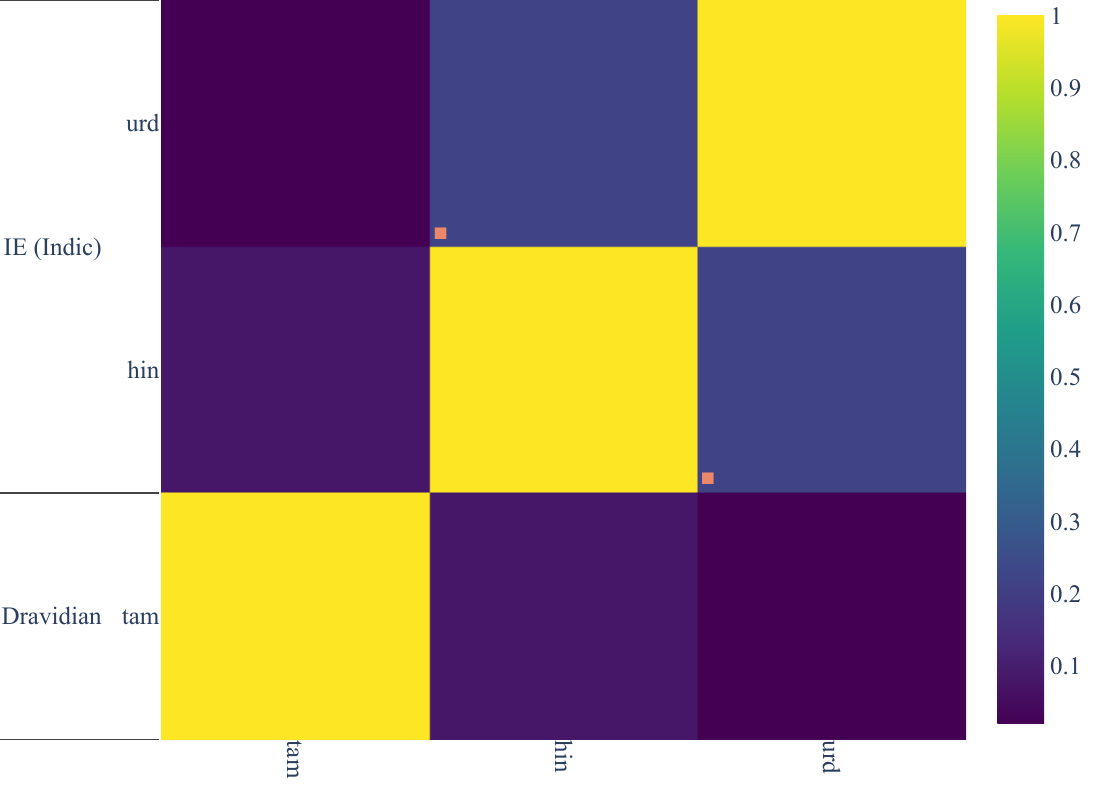}
         \caption{Finiteness}
     \end{subfigure}    
    \newline
     \begin{subfigure}[b]{0.23\textwidth}
         \centering
         \includegraphics[width=\textwidth]{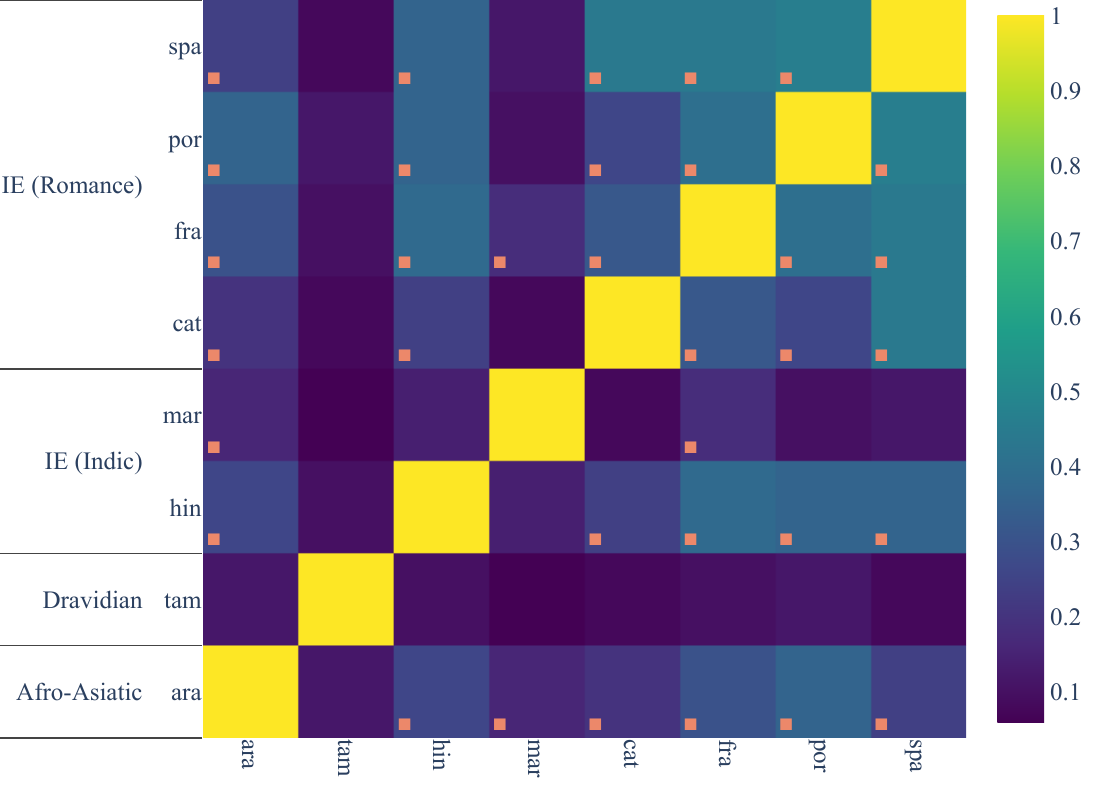}
         \caption{Gender}
     \end{subfigure}
     \hfill
     \begin{subfigure}[b]{0.23\textwidth}
         \centering
         \includegraphics[width=\textwidth]{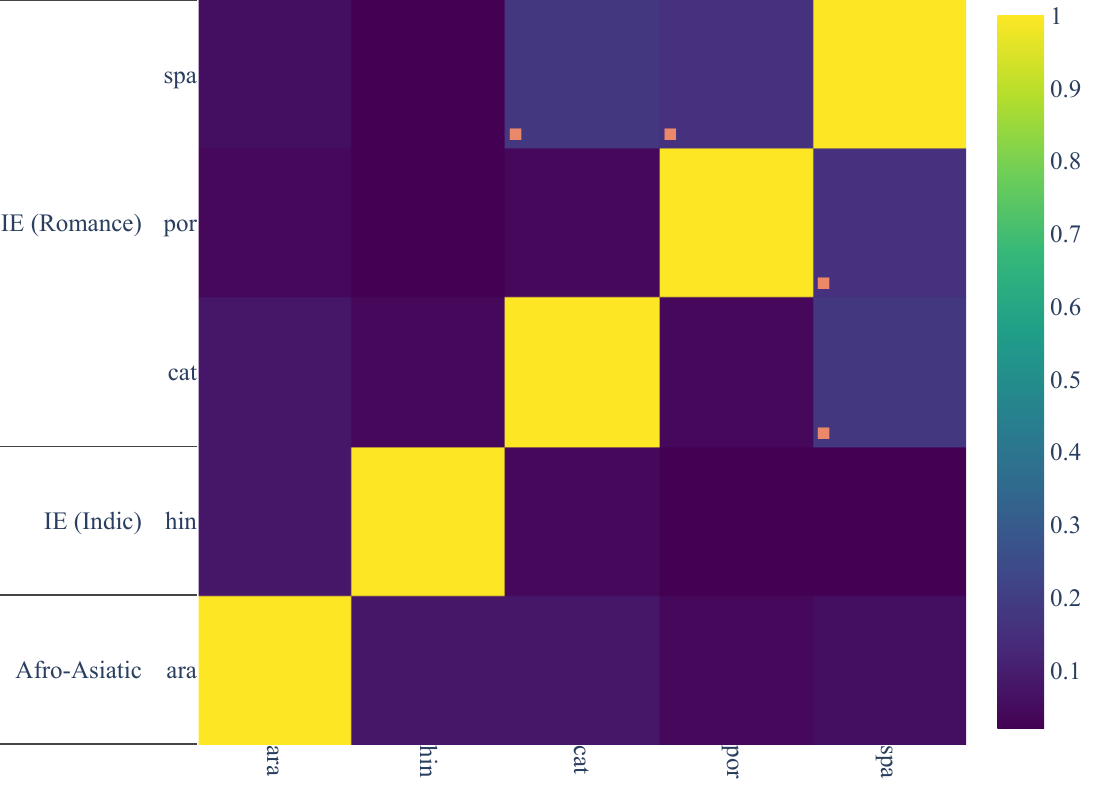}
         \caption{Mood}
     \end{subfigure}  
     \begin{subfigure}[b]{0.23\textwidth}
         \centering
         \includegraphics[width=\textwidth]{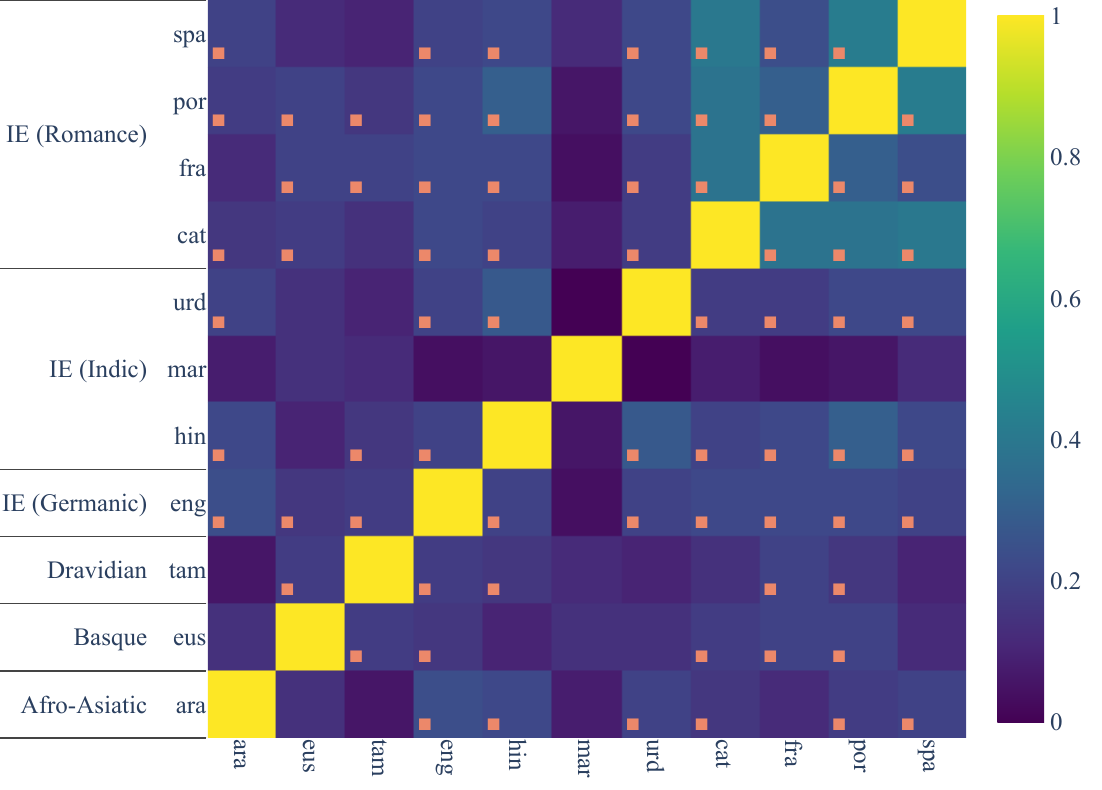}
         \caption{Number}
     \end{subfigure}
     \hfill
     \begin{subfigure}[b]{0.23\textwidth}
         \centering
         \includegraphics[width=\textwidth]{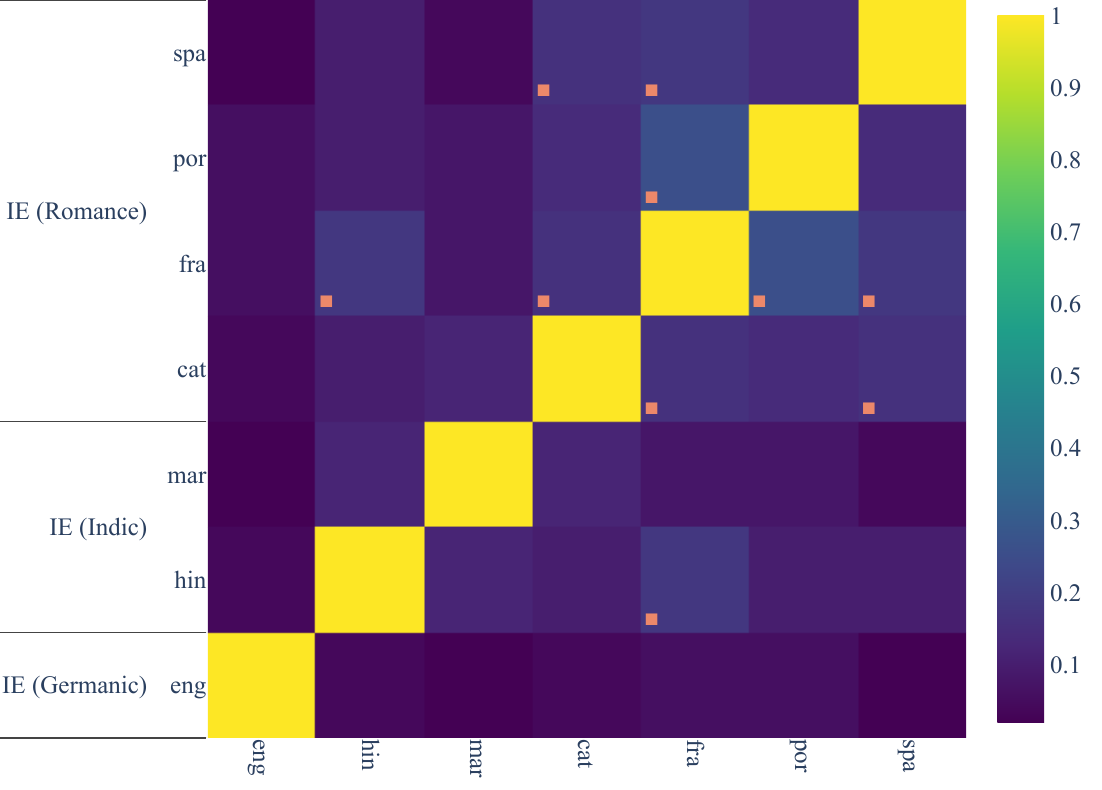}
         \caption{Person}
     \end{subfigure}  
        \newline
     \begin{subfigure}[b]{0.23\textwidth}
         \includegraphics[width=\textwidth]{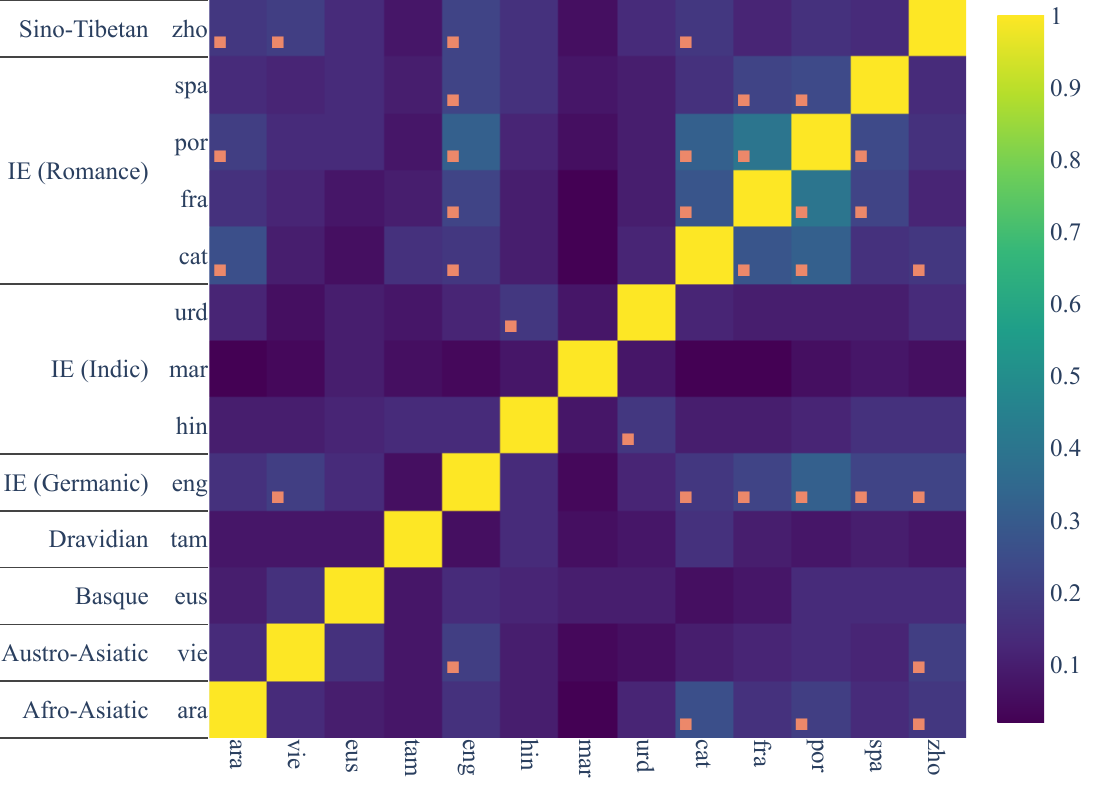}
         \caption{POS}
     \end{subfigure}
     \hfill
     \begin{subfigure}[b]{0.23\textwidth}
         \includegraphics[width=\textwidth]{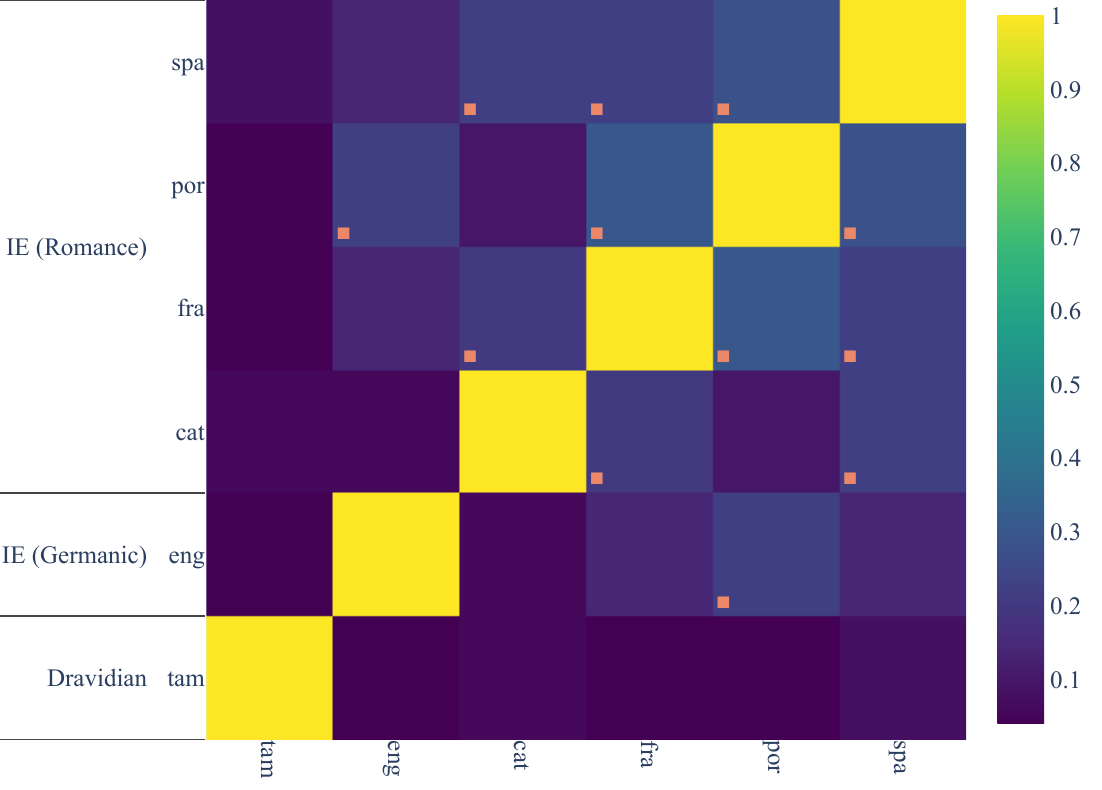}
         \caption{Tense}
     \end{subfigure} 
    \hfill
     \begin{subfigure}[b]{0.23\textwidth}
         \includegraphics[width=\textwidth]{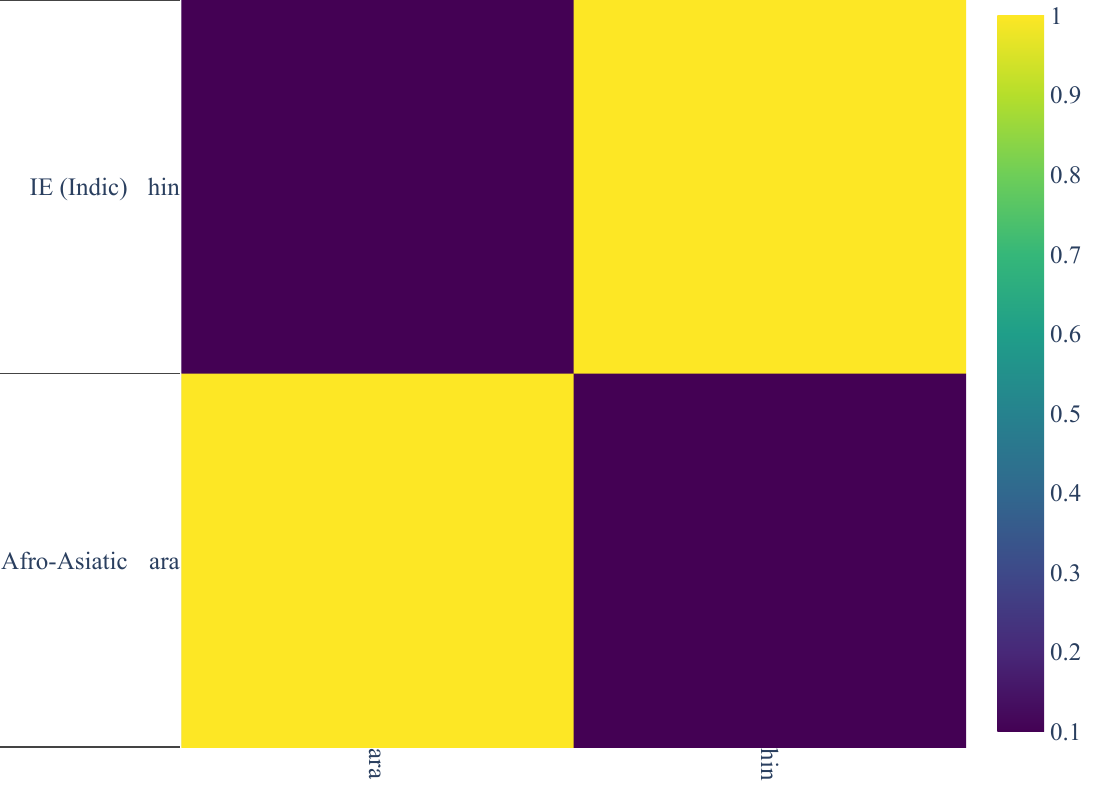}
         \caption{Voice}
     \end{subfigure}
\end{figure}

\clearpage

\subsection{Pairwise overlap rates throughout training}
The neuron overlap rate between target languages and English in Layer 13: 
\begin{figure}[H]
    \centering
    \begin{subfigure}[b]{0.3\textwidth}
        \centering
        \includegraphics[width=\textwidth]{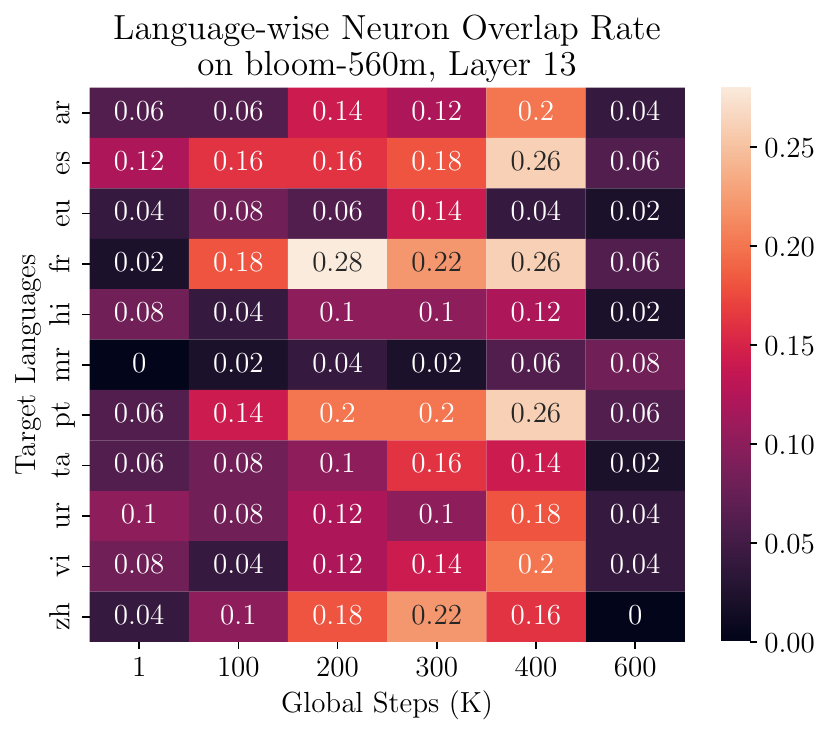}
        \caption{bloom-560m}
    \end{subfigure}
    \begin{subfigure}[b]{0.3\textwidth}
        \centering
        \includegraphics[width=\textwidth]{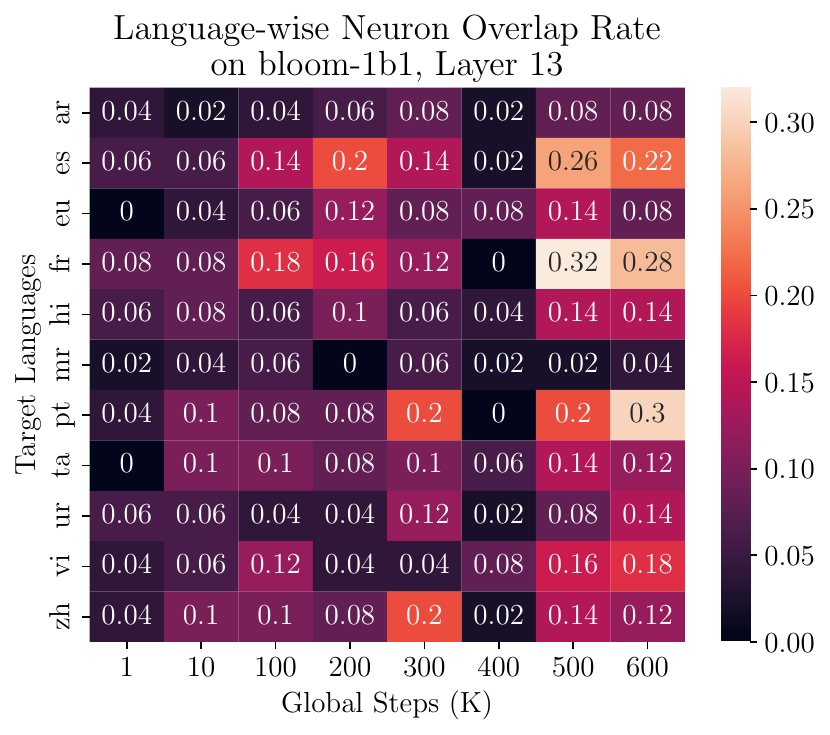}
        \caption{bloom-1b1}
    \end{subfigure}
    \begin{subfigure}[b]{0.3\textwidth}
        \centering
        \includegraphics[width=\textwidth]{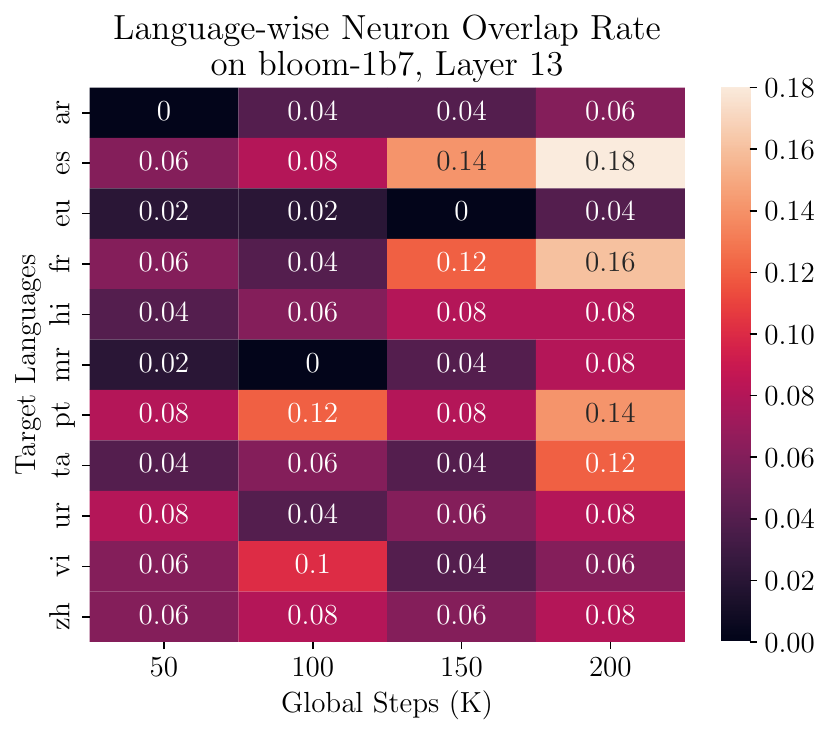}
        \caption{bloom-1b7}
    \end{subfigure}
\end{figure}
The neuron overlap rate between target languages and English in Layer 17: 
\begin{figure}[H]
    \centering
    \begin{subfigure}[b]{0.3\textwidth}
        \centering
        \includegraphics[width=\textwidth]{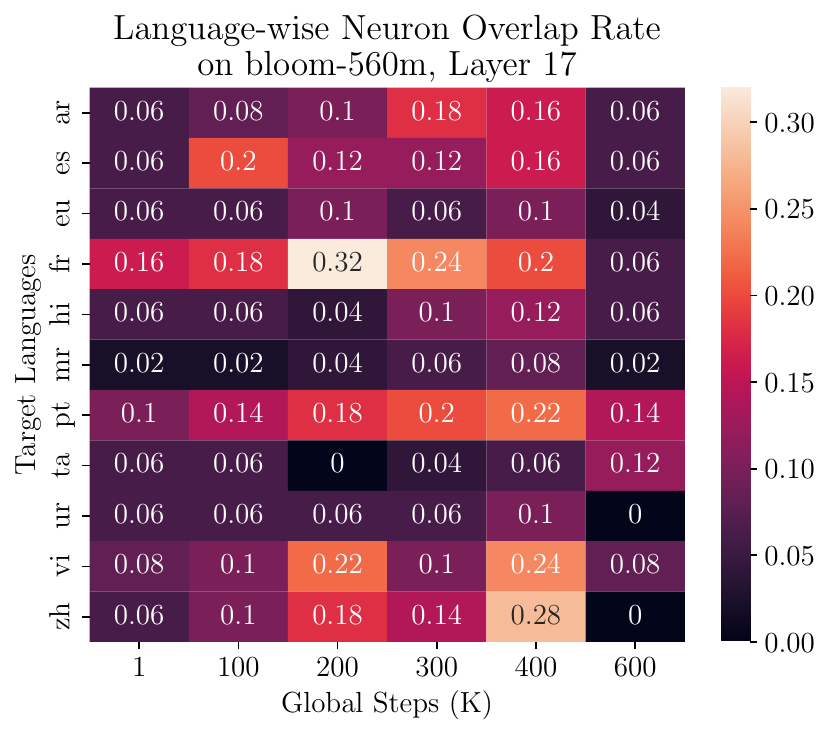}
        \caption{bloom-560m}
    \end{subfigure}
    \begin{subfigure}[b]{0.3\textwidth}
        \centering
        \includegraphics[width=\textwidth]{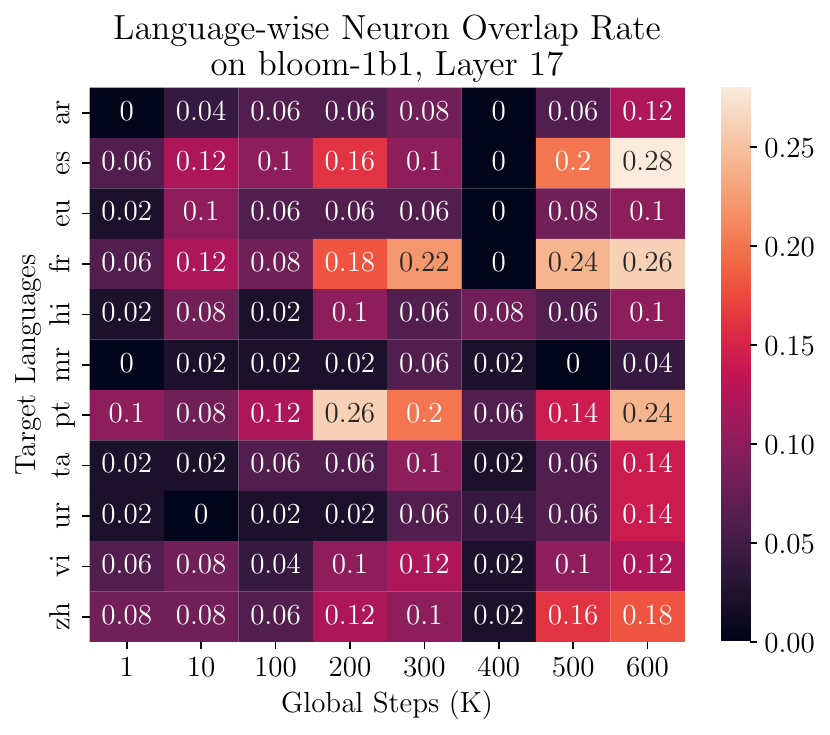}
        \caption{bloom-1b1}
    \end{subfigure}
    \begin{subfigure}[b]{0.3\textwidth}
        \centering
        \includegraphics[width=\textwidth]{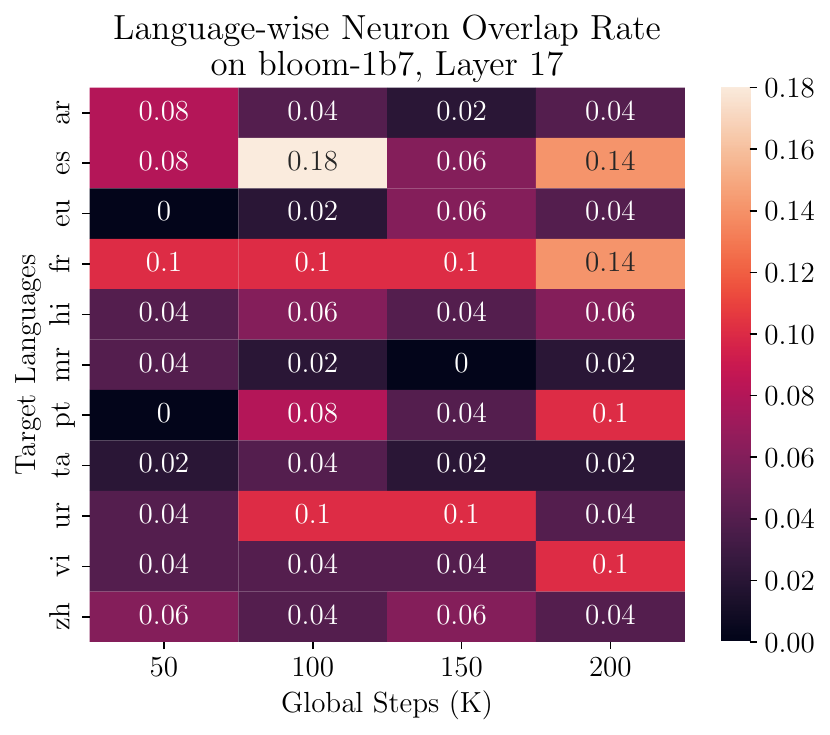}
        \caption{bloom-1b7}
    \end{subfigure}
\end{figure}

\subsection{Zero-shot Cross-lingual Performance on Downstream Tasks}
\label{apdx:zero_shot_performance}
\begin{figure}[H]
    \centering
    \begin{subfigure}[b]{0.3\textwidth}
        \centering
        \includegraphics[width=\textwidth]{Plots/heatmaps/bloom-560m_xnli_acc.pdf}
        \caption{bloom-560m, XNLI}
    \end{subfigure}
    \begin{subfigure}[b]{0.3\textwidth}
        \centering
        \includegraphics[width=\textwidth]{Plots/heatmaps/bloom-1b1_xnli_acc.pdf}
        \caption{bloom-1b1, XNLI}
    \end{subfigure}
    \begin{subfigure}[b]{0.3\textwidth}
        \centering
        \includegraphics[width=\textwidth]{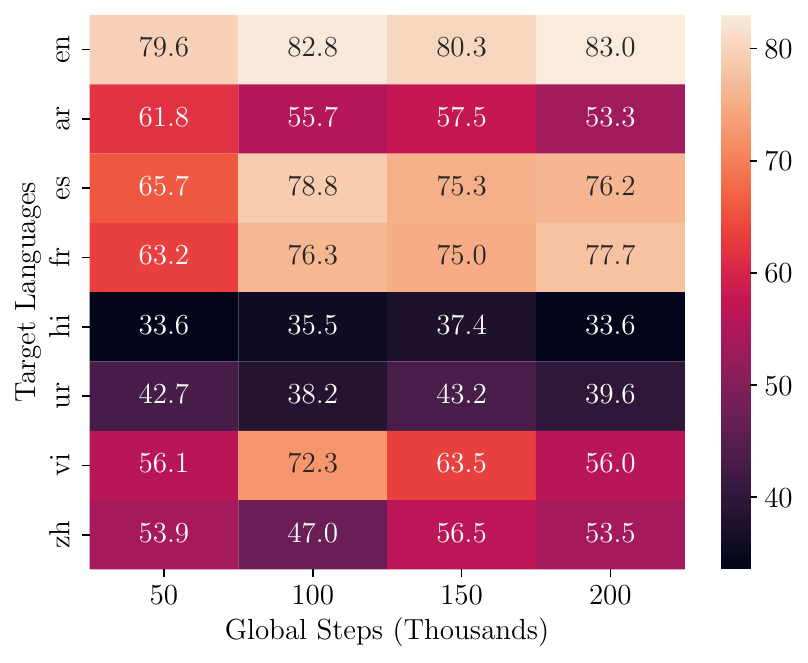}
        \caption{bloom-1b7, XNLI}
    \end{subfigure}
\end{figure}

\begin{figure}[H]
    \centering
    \begin{subfigure}[b]{0.3\textwidth}
        \centering
        \includegraphics[width=\textwidth]{Plots/heatmaps/bloom-560m_f1_score.pdf}
        \caption{bloom-560m, POS tagging}
    \end{subfigure}
    \begin{subfigure}[b]{0.3\textwidth}
        \centering
        \includegraphics[width=\textwidth]{Plots/heatmaps/bloom-1b1_f1_score.pdf}
        \caption{bloom-1b1, POS tagging}
    \end{subfigure}
    \begin{subfigure}[b]{0.3\textwidth}
        \centering
        \includegraphics[width=\textwidth]{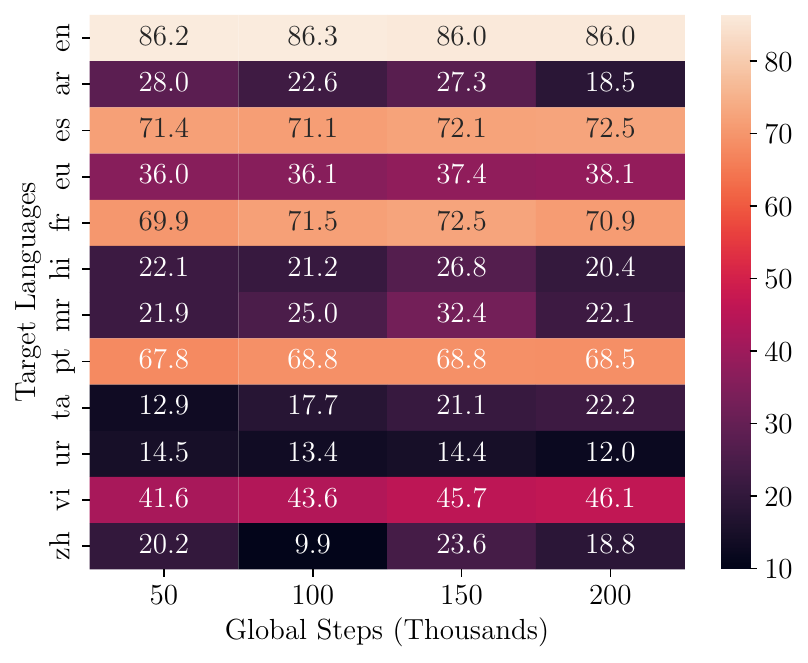}
        \caption{bloom-1b7, POS tagging}
    \end{subfigure}
\end{figure}

\end{document}